# Integrating Vehicle Slip and Yaw in Overarching Multi-Tiered Automated Vehicle Steering Control to Balance Path Following Accuracy, Gracefulness, and Safety

Ming Xin and Mark A. Minor
Department of Mechanical Engineering
University of Utah, Salt Lake City, UT
corresponding author: ming.xin@outlook.com

*Abstract*— Balancing path following accuracy and error convergence with graceful motion in steering control is challenging due to the competing nature of these requirements, especially across a range of operating speeds and conditions. This paper demonstrates that an integrated multi-tiered steering controller considering the impact of slip on kinematic control, dynamic control, and steering actuator rate commands achieves accurate and graceful path following. This work is founded on multi-tiered sideslip and yaw-based models, which allow derivation of controllers considering error due to sideslip and the mapping between steering commands and graceful lateral motion. Observer based sideslip estimates are combined with heading error in the kinematic controller to provide feedforward slip compensation. Path following error is compensated by a continuous Variable Structure Controller (VSC) using speed-based path manifolds to balance graceful motion and error convergence. Resulting yaw rate commands are used by a backstepping dynamic controller to generate steering rate commands. A High Gain Observer (HGO) estimates sideslip and yaw rate for output feedback control. Stability analysis of the output feedback controller is provided, and peaking is resolved. The work focuses on lateral control alone so that the steering controller can be combined with other speed controllers. Field results provide comparisons to related approaches demonstrating gracefulness and accuracy in different complex scenarios with varied weather conditions and perturbations.

## I. INTRODUCTION

Robotic vehicles offer potential for improving transportation safety and efficiency, but present new challenges. While this is a multifaceted field, a major researcher area is vehicle steering control [1]. A few examples include accident avoidance systems [2], Automated Ground Vehicles (AGV) [3-10], and Automated Highway Systems (AHS) [11, 12]. All of these require vehicle steering control, which is the subject of this paper. While steering has been considered for decades, achieving accurate ground vehicle path following presents challenges. Speeds vary significantly during operation, which requires that kinematic and dynamic controllers consider how speed and steering affect lateral acceleration and sideslip. Sideslip reduces tracking accuracy, which depends on surface conditions, environment, and tire properties. Mass distribution varies, which affects sideslip at the front and rear tires, causing vehicle steering response to vary. Disturbances from wind, road cant and slope, and path curvature variation further effect sideslip and path following error.

As a result, robotic transportation systems present challenges with sideslip, uncertainty, and disturbances. At the same time, steering systems need to provide safe and graceful motion. *Safe* steering commands result in lateral acceleration below physics-based thresholds whereas *gracefulness* refers to lateral motion with minimal oscillations and smaller magnitude. Traditional robot controllers are not suited to this. The classic paradigm of tiered kinematic and dynamic controllers is limited since these approaches assume that kinematics are decoupled from dynamics [8, 13, 14]. This allows the kinematic and dynamic controllers to be derived separately, but sideslip caused by dynamics affects the kinematics and must be considered [15], which is the focus of this paper. Lateral control is the primary focus of the work such that the steering controller can easily be combined with higher level motion planners and separate speed controllers.

### A. Proposed Framework

Building upon our prior work [10] and [16], the proposed steering controller approach uses a slip-coupled multi-tiered kinematic-dynamic framework, Fig 1, to consider sideslip, $\beta$. This allows the kinematic controller to anticipate sideslip. Yaw rate, $r_{kin}$, is output from the kinematic controller to the dynamic controller, which considers slip, yaw rate, and actuator response.

A slip-yaw dynamic model [17] is selected since yaw rate naturally leads to centripetal forces and sideslip. Yaw rate is easily sensed by gyroscopes such that the dynamic controller can match sensing and command data. This is opposed to the typical robotics approach where configuration space variables describe desired robot motion and the dynamic controller steers the robot to reduce error [13, 14]. Configuration space variables calculate path following error in kinematics, but since GPS is noisy, that information is filtered and fused with other sensors to reduce noise [18]. While filters can reduce noise, the effect of noise on dynamic variables, such as yaw rate and acceleration, is still notable. Hence the dynamic controller uses yaw rate sensing directly, which allows direct comparison of input commands and sensor output. The output of the dynamic controller is steering rate, $\hat{\omega}$.

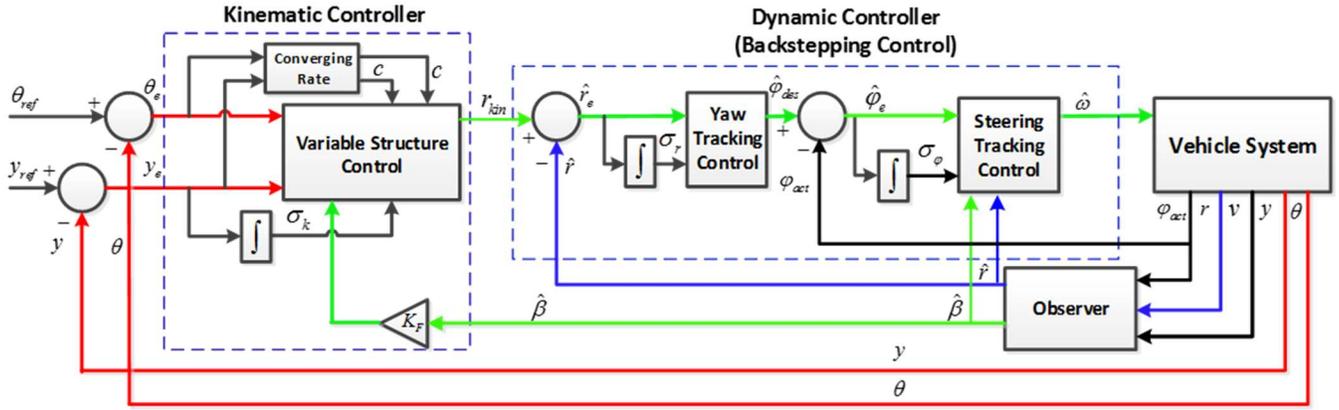

Fig 1: The Over-arching Structure of Vehicle Steering Controller. For symbol definitions refer to Sec II and III.

A slip-yaw observer estimates sideslip and yaw rate for output feedback control. Sideslip is difficult to measure, and yaw-rate derivatives are required in the controller, but they are noisy. A high-gain observer reduces noise and compensates for uncertainty, but does require special consideration to avoid peaking [19].

### B. Proposed Controllers

The kinematic controller is designed to consider sideslip and provide yaw rate commands that lead to accurate and graceful motion. Continuous Variable Structure Control (VSC) modifies driving behavior as a function of vehicle speed and path following error where $tanh()$ functions are used to resolve chatter while providing smooth robust commands [20]. Hierarchal path manifolds balance convergence of heading error and lateral error to provide graceful motion. Sideslip is compensated by body slip terms in the controller. Errors in slip and the model are treated as uncertainties and compensated by continuous robust terms. The integral of lateral error further compensates uncertainty and disturbances. Saturation is applied to the kinematic controller output since the combination of robust controllers and observers are known to create peaking [19]. *The result is graceful and robust kinematic control designed.*

Dynamic control uses backstepping to couple desired yaw rate, slip-yaw dynamics, and actuator rate for improved response. A yaw-rate tracking controller is designed based upon the slip-yaw model, which results in steering angle commands. Dynamic extension and backstepping allow steering rate to serve as the input to the steering system, allowing terms from the desired path and kinematic controller to map to steering rate. Integrators compensate for uncertainty in yaw rate, steering map, and disturbances.

Only lateral steering control is considered in this work, however, meaning that separate longitudinal controls must maintain reasonable speeds and accelerations. A human provides speed control here.

### C. Evaluations

Validation is provided by field experiments in challenging scenarios at varying speeds. Paths provide varying complexity on different surfaces with disturbances from slopes and bumps in rainy and dry conditions. Metrics quantify tracking error and graceful motion. Comparisons between the proposed and baseline controllers highlight the benefits of the proposed methodology.

### D. Structure of Paper

Sec. II reviews related work and describes contributions. Sec III presents kinematic and dynamic models. The kinematic controller, path manifold, and dynamic controller are presented in Secs IV, V, and VI. Observer design and output feedback control are in Sec VII. Sec VIII provides experimental procedures. Sec IX presents results with discussion in Sec X. Conclusions are in Sec XI.

## II. RELATED WORK

Self-driving cars have been of keen interest for decades. The 1939 World's Fair portrayed cities connected by highways with automatically driven cars, leading to radio controlled vehicles in the '50's [21]. The California Partners for Advanced Transit and Highways (PATH) project [11, 12, 22] led to Demonstration '97 using magnetic beacons. The DARPA Grand Challenge (DGC) [4] followed with sensor based navigation of dirt roads. The DARPA Urban Challenge (DUC) [3, 5, 6] examined operation in cities. Industry has focused on sensor-based navigation. Google Car emphasized navigation on freeways in '09 [23]. Tesla deployed automated steering and navigation [24]. Industry is not publishing, but patent disclosures indicate emphasis on sensor-based navigation [25-28]. Like the DUC [3, 5, 6] the emphasis is on path planning assuming slip can be compensated by other methods.

Slip determines vehicle/ground dynamic interactions and should be considered to allow accurate and graceful motion. PATH [11, 12, 22] used lateral dynamics to design a steering controller balancing tracking accuracy and passenger comfort. They used gain-scheduling to balance trade-offs, but their magnetic beacon localization does not scale to real application. A few researchers considered physical effects during the DGC and DUC such as steering actuator dynamics [3], tire-slip [4], and velocity variation [6]. [4] relied on kinematic control but considered tire slip and cornering stiffness to determine steering angles commands, but neither graceful motion nor active control of vehicle dynamics are described. [3] considered steering actuator dynamics and used discretized steering rate commands, but this can create "jerky" motion. After the DUC, [7] provided tighter integration using a longitudinal dynamic model with a standard non-slip bicycle kinematic model. Tracking improved, but speeds and speed variations are small; neither actuator dynamics nor sideslip are considered directly.

Some consider the impact of dynamics on steering control. [29] use a bicycle model to design robust yaw stability control, but do not consider kinematics. [30] and [31] work on coupled longitudinal and lateral dynamic controls. They introduce a tire model for better control; the kinematic model generates references for dynamic controllers, but the impact of kinematics on steering control is neglected. [29-31] provide simulations to verify their work, lacking physical validation.

A few researchers consider kinematic and dynamic models similar to ours. [8] applies an open-loop tiered controller to the tire-ground dynamic model. Only simulations are provided and controllers require burdensome instrumentation. [9] uses an architecture and slip-yaw model similar to here with a computationally demanding predictive controller leading to very low speeds, ignoring sideslip and steering dynamics. [32] linearizes models for feedforward look-ahead steering with classic state feedback control. [33] uses a slip-based kinematic model with a sideslip observer and adaptive controller with sideslip feedback compensation. [34] linearized models for model predictive control for high-level yaw moment control and low-level PI force control with prohibitive embedded load cells on a non-standard test platform. This paper fixes deficiencies in [8] and [9] and achieves better performance under varying speeds and in complex environmental conditions with limited sensing. Different from [32-34], this work uses nonlinear slip-based kinematics to handle perturbations and distributes error convergence in four nested tracking controllers.

This paper expands conference papers [10, 16, 35-37] to provide archival presentation of matured algorithms, expanded analysis, and rigorous experimental evaluations not presented before. SMC was based on a nonslip kinematic model in [35, 36]. [35] used a simple dynamic controller and [36] added backstepping to improve actuator control. Continuous VSC replaced continuous SMC to deal with varying speeds and slip [10]. [16] added integrators for lateral error and the difference between the actual heading error and that desired by the continuous VSC controller, which was named "sideslip disturbance", but this is not entirely accurate since it is not purely due to sideslip. *This paper uses a slip-based kinematic model and expands derivation of the continuous VSC controller to better consider sideslip compensation, improved methods of adapting the controller online, and physical limitations.* This work examines the relationship between rear-slip and sideslip, demonstrating that sideslip compensation improves stability and gracefulness. The kinematic controller uses the integral of lateral error, but not "sideslip disturbance" like [16] since this paper compensates sideslip directly. *Thus, slip compensation in the proposed kinematic controller is improved.*

Like [10, 16, 36, 37], this paper uses backstepping in dynamics to improve steering commands. [10, 36] produce *steering angle* commands whereas [16, 37] provided *steering rate* commands for better tracking, which is used here. A high-gain observer is applied as in [10, 16, 35, 36], but this work and [37] apply saturation [38] to limit yaw-rate commands from the kinematic controller, reducing "peaking". Sideslip feedback to the kinematic controller also reduces demand on the kinematic controller and peaking. *More graceful motion results, which is especially notable with rapid path curvature variations.* Evaluations are expanded, demonstrating improved tracking and more graceful motion, even with challenging environments. This work is derived from the dissertation [39].

## III. MODELING

Bicycle models are the basis of this research due to their simplicity and sufficiency for modelling the relationships between steering commands, trajectory tracking, and vehicle dynamics. A tire-slip based kinematic model is developed first, followed by the slip-yaw dynamic model and model coupling.

### A. Slip-Based Kinematic Models

Fig 2 presents a slip-based bicycle model of the vehicle kinematics in the global reference frame $\{O_0\}$. Coordinate frame $\{O_B\}$ at the rear axle represents the actual posture. The frame is rotated by heading angle $\theta$ such that the $x_B$ axis is aligned with the *"longitudinal direction"* of the vehicle and the $y_B$ axis is the *"lateral direction"*. $O_B$ is located by $\boldsymbol{R_B} = [x_{act}, y_{act}]^T$ in the global frame $\{O_0\}$ such that the vehicle actual posture is $\boldsymbol{P_{act}} = [x_{act} \quad y_{act} \quad \theta]^T$. The front and rear tire velocities at A and $O_B$ are expressed in $\{O_0\}$ as $\boldsymbol{v_A}$ and $\boldsymbol{v_B}$, respectively. Front slip angle, $\alpha_f$, and rear slip angle, $\alpha_r$, are measured relative to the vehicle in $\{O_B\}$, where steering angle is $\varphi$. Slip angles are shown in positive directions, but a positive steering angle $\varphi$ usually results in negative slip angles. The slip-based kinematic state model is then:

$$\dot{x}_{act} = v_B \cos(\theta + \alpha_r) \qquad (1)$$
$$\dot{y}_{act} = v_B \sin(\theta + \alpha_r) \qquad (2)$$

where $v_B$ is the magnitude of $\boldsymbol{v_B}$. The $\theta$ state equation is derived noting that the vehicle is rotating about its instantaneous center of rotation (ICR) $o'$ at the yaw rate $r$ in Fig 3. Thus, $\dot{\theta} = r = v_B/\overline{o'B}$, is expressed in terms of rear tire slip and steering angles,

$$\dot{\theta} = r = \left(\frac{v_B}{L}\right)\cos(\alpha_r)(\tan \alpha_r + \tan(\varphi + \alpha_f)) \qquad (3)$$

where steering angle $\varphi$ is based upon steering rate $\omega$,

$$\dot{\varphi} = \omega. \qquad (4)$$

Coordinate frame $\{O_D\}$ is the desired posture on the path located by $\boldsymbol{R_D} = [x_{ref}, y_{ref}]^T$. $\{O_D\}$ is tangential to the path and desired heading $\theta_{ref}$ describes the orientation in $\{O_o\}$. The reference posture is then $\boldsymbol{P_{ref}} = [x_{ref} \quad y_{ref} \quad \theta_{ref}]^T$ and the nonslip reference model is,

$$\dot{x}_{ref} = v_{ref} \cos \theta_{ref} \qquad (5)$$
$$\dot{y}_{ref} = v_{ref} \sin \theta_{ref} \qquad (6)$$
$$\dot{\theta}_{ref} = \kappa_{ref} v_{ref} \qquad (7)$$

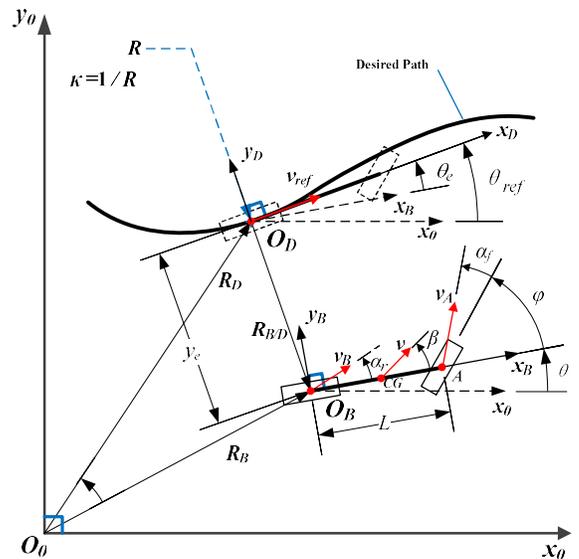

Fig 2: Vehicle Kinematic Model

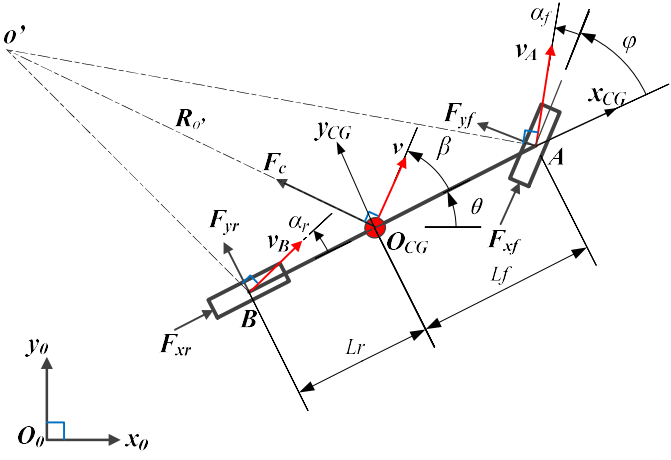
Fig 3. Slip-Yaw Vehicle Dynamics Model

where $v_{ref}$ is the magnitude of $\mathbf{v_{ref}}$ and path curvature at $O_D$ is $\kappa_{ref} = 1/R_{ref}$, where the reference turning radius is $R_{ref}$ and yaw rate is $\dot{\theta}_{ref} = r_{ref} = v_{ref}/R_{ref}$ such that $\dot{\theta}_{ref} = \kappa_{ref} v_{ref}$.

Path following error state equations are then formulated in $\{O_D\}$, where the distance between $O_D$ and $O_B$ along the $\mathbf{x_D}$ axis is the longitudinal error, $x_e$, and the distance along the $\mathbf{y_D}$ axis is the lateral error, $y_e$. The angle between $\mathbf{x_B}$ and $\mathbf{x_D}$ is the heading error, $\theta_e$. $\dot{\theta}_e$ is the time derivative of $\theta_e = \theta_{ref} - \theta$. Position error states are derived by expressing tracking error in $\{O_0\}$,

$$\begin{bmatrix} {}^0x_e \\ {}^0y_e \end{bmatrix} = \begin{bmatrix} x_{ref} - x_{act} \\ y_{ref} - y_{act} \end{bmatrix} \quad (8)$$

Tracking error (8) is represented in $\{O_D\}$ via rotation as $x_e$ and $y_e$, Fig 2. Heading error is defined as $\theta_e = \theta_{ref} - \theta$. Time derivatives are applied and (1), (2), (5), and (6) are substituted. Trigonometric transformations produce the *Slip-Based Kinematic Tracking Error Model* in the $\{O_D\}$ coordinate frame,

$$\dot{x}_e = v_{ref} - v_B \cos(\theta_e - \alpha_r) + y_e \dot{\theta}_{ref} \quad (9)$$
$$\dot{y}_e = v_B \sin(\theta_e - \alpha_r) - x_e \dot{\theta}_{ref} \quad (10)$$
$$\dot{\theta}_e = \dot{\theta}_{ref} - \dot{\theta} = \kappa_{ref} v_{ref} - r. \quad (11)$$

**Proposition 1:** The initial posture on the reference path is selected such that $x_e(0) = 0$. $v_{ref}$ is calculated based upon the actual vehicle speed such that $\dot{x}_e = 0$, thus $x_e \triangleq 0$. (9) drops out and (10) becomes (12). (11) is restated as (13) for completeness:

$$\dot{y}_e = v_B \sin(\theta_e - \alpha_r) \quad (12)$$
$$\dot{\theta}_e = \kappa_{ref} v_{ref} - r. \quad (13)$$

This is the *Slip-Based Kinematic Path Following Error Model*, which is the basis of the controller design. ∎

Proposition 1 simplifies design of the kinematic controller since only lateral error and heading error are compensated. The addition of sideslip can be appreciated [10, 35, 36] by noting rear tire slip angle term $\alpha_r$ in (12).

### B. Slip-Yaw Dynamics

Dynamics are characterized by a linear slip-yaw bicycle model [40] that considers the interaction of vehicle inertia, tire-ground interaction, and steering, Fig 3. Motion is described by (1)-(4), whereas this section presents dynamic state equations for slip angle and yaw rate. They are presented briefly since they are standard [40, 41], but form the basis of the dynamic controller.

Coordinate frame $\{O_{CG}\}$ is attached to the Center of Gravity (CG), such that the $x_{CG}$ axis is along the longitudinal axis of the vehicle and $y_{CG}$ is along the lateral direction, similar to $\{O_B\}$, except $O_{CG}$ is at the CG along the $x_B$ axis. $L_f$ and $L_r$ are the distances between the CG and the front and rear axles. $L = L_f + L_r$ is the vehicle wheelbase. In $\{O_0\}$, the vehicle velocity vector is $\mathbf{v}$ (with magnitude $v$), which has sideslip, $\beta$, measured relative to the $x_{CG}$ axis. Heading angle is $\theta$ and its time derivative is yaw rate, $r$; steering angle and rate are again $\varphi$ and $\omega$.

Typical assumptions are made to derive the linear slip-yaw dynamic model [40, 41], but we later demonstrate that perturbations resulting from violating the assumptions can be compensated:

Assumption 1: Ackerman steering is assumed such that steering angle, $\varphi$, describes the effective instantaneous center of rotation, $o'$, created by front wheels of the vehicle, Fig 3.

Assumption 2: Lateral and longitudinal dynamics can be separated and coupling effects can be ignored.

Assumption 3: *Typical driving scenarios* are assumed: ($a$) *Moderate to high speeds* ranging from 1 $m/s$ to 40 $m/s$, ($b$) *Limited Speed Variations* are assumed for deriving the dynamic model, but perturbations due to violating this are compensated by the high gain observer, ($c$) *Limited Steering Angle* is required to drive; $|\varphi| < 11°$ is typical, ($d$) *Limited Road Curvatures* are based upon limited steering angle $\varphi$, where $|\kappa| \leq \tan(11°)/L = 0.0658\ m^{-1}$ for the vehicle described in Sec VIII(A), ($e$) *Varying Road Surfaces*: concrete, asphalt, and gravel, ($f$) *Varying Road Conditions*: dry or rainy/wet, ($g$) *Road Side Slope and Incline* are assumed flat and level to simplify derivations, although results show good performance on sloped terrain due to the HGO. Variations in speed and terrain are compensated as disturbances.

Assumption 4: *Limited Slip Angles*, $|\alpha_f| < 11°, |\alpha_r| < 11°$, and, $|\beta| < 11°$ follow operating conditions in Assumption 3.

Assumption 5: *In-plane Rigid Body Motion* is assumed such that body roll due to vehicle suspension is neglected. Wheel forces are lumped into single wheels modelled at the center of the axles.

**Proposition 2**: Given the above assumptions, the *Slip-Yaw Dynamic Model* of the vehicle can be approximated as,

$$\dot{\beta} = a_{11}\beta + a_{12}r + b_{11}\varphi + \delta_\beta \triangleq \phi_\beta(\beta, r, \varphi) \quad (14)$$
$$\dot{r} = a_{21}\beta + a_{22}r + b_{21}\varphi + \delta_r \triangleq \phi_r(\beta, r, \varphi) \quad (15)$$
$$\dot{\varphi} = \omega \quad (16)$$

where $\beta, r$, and $\varphi$ are the vehicle slip angle, yaw rate, and steering angle, $a_{11} = -\frac{(C_f + C_r)}{mv}$, $a_{12} = -\left(1 + \frac{C_f L_f - C_r L_r}{mv^2}\right)$, $a_{21} = -\frac{(C_f L_f - C_r L_r)}{J}$, $a_{22} = -\frac{(C_f L_f^2 + C_r L_r^2)}{Jv}$, $b_{11} = \frac{C_f}{mv}$, $b_{21} = \frac{C_f L_f}{J}$, and $\omega$ is the steering rate. $C_f$ and $C_r$ are the front and rear cornering stiffnesses, $m$ and $J$ are the vehicle mass and rotational inertia. $\delta_\beta$ and $\delta_r$ are the higher order terms resulting from linearization of the nonlinear model and perturbations due to speed variations and terrain, which are not considered in the literature [40, 41].

Proof: Per Assumption 4 and Fig 3, planar rigid body dynamics characterize forces acting on the vehicle and its resulting motion in the lateral and yaw directions, respectively, as,

$$F_{yf}\cos\varphi + F_{yr} + F_{xf}\sin\varphi = F_c \cos\beta + m\dot{v}\sin\beta \quad (17)$$
$$F_{xf}\sin\varphi\ L_f + F_{yf}\cos\varphi\ L_f - F_{yr} L_r = J\ddot{\theta} = J\dot{r} \quad (18)$$

where it is noted that yaw acceleration $\dot{r} = \ddot{\theta}$. Tire forces are characterized by longitudinal and lateral components expressed relative to each tire; $F_{xf}$ and $F_{xr}$ are front and rear tire longitudinal forces, respectively, whereas $F_{yf}$ and $F_{yr}$ are lateral forces. Centripetal force $F_c$ at the CG is directed toward the ICR, $o'$, which can be expressed as $F_c = mv^2/R$, where $R$ is the turning radius. Noting that the angular rate of the vehicle can be expressed as $(v/R) = (\dot{\theta} + \dot{\beta})$, the result is $F_c = mv(r + \dot{\beta})$, which is substituted into (17) to produce the sideslip state equation (14).

Per Assumption 3 (b), braking and acceleration forces are small during steering such that $F_{yf} \gg F_{xf}, F_{yr} \gg F_{xr}$. Hence $F_{xf}, F_{xr}$, and $\dot{v}$ are neglected. Perturbations due to acceleration and deceleration are lumped into $\delta_\beta$. Lateral tire forces are approximated by the cornering stiffness model, $F_{yf} \cong -\mu_f C_{f0}\alpha_f = -C_f\alpha_f$, and $F_{yr} \cong -\mu_r C_{r0}\alpha_r = -C_r\alpha_r$. $\mu_f$ and $\mu_r$ are front and rear tire friction coefficients, $C_{f0}$ and $C_{r0}$ are *tire cornering stiffnesses*, $C_f$ and $C_r$ are *cornering stiffnesses,* and $\alpha_f$ and $\alpha_r$ are front and rear tire slip angles. These terms are combined with the equations of motion to produce (14)-(15). ∎

Zero velocity is problematic due to $v$ in the denominators of the coefficients in (14)-(15). Substituting a small positive velocity $v_\epsilon$, when $v \approx 0$ alleviates numerical issues. Sensor noise affects the magnitude of $v_\epsilon$ that is selected.

### C. Resolving Model Reference Frames

One challenge is that $O_B$ and $O_{CG}$ are at different locations. Slip angles $\beta$ and $\alpha_r$ and speeds in the different reference frames are resolved here. Analysis of slip angles begins with geometric analysis of the vehicle and velocity headings relative to the instantaneous center of rotation, $o'$, Fig 3. In the triangle $\triangle o' O_{CG} B$, the Law of Sines is applied to show $\frac{\sin(\angle Bo'O_{CG})}{L_r} = \frac{\sin(\angle o'BO_{CG})}{R}$, which is equivalent to $\frac{\sin(\beta-\alpha_r)}{L_r} = \frac{\sin(90°+\alpha_r)}{R} = \frac{\cos(-\alpha_r)}{R}$. Based on small angle assumptions per Assumption 4, $\sin(\beta - \alpha_r) \approx \beta - \alpha_r$, $\cos(-\alpha_r) \approx 1$, which is applied to the previous equations to show

$$\frac{\beta-\alpha_r}{L_r} = \frac{1}{R} = \kappa. \tag{19}$$

Based on the kinetic analysis, $\Sigma M_A = -F_{yr}L + F_c \cos\beta\, L_f = 0$ in steady cornering, where $F_{yr} = -C_r\alpha_r$ and $F_c = \frac{mv^2}{R} = mv^2\kappa = mv^2\frac{\beta-\alpha_r}{L_r}$. Combining equations results in $C_r\alpha_r L + \frac{mv^2 L_f}{L_r}(\beta - \alpha_r) = 0$, which can be reduced to:

$$\alpha_r \cong \frac{-1}{\left(\frac{C_r L L_r}{mv^2 L_f} - 1\right)}\beta. \tag{20}$$

This expression will serve as a baseline for understanding the effect of sideslip compensation proposed in the next section.

**Proposition 3**: *Speed approximation:* The rear axle speed, $v_B$, and the CG speed, $v$, are approximately equal, i.e. $v_B \approx v$. ∎

Proposition 3 is appreciated by considering that the longitudinal components of $v_B$ and the CG velocity, $v$, must be equal. As a result, $v_B \cos\alpha_r = v \cos\beta$. According to Assumption 4, $\cos\alpha_r \approx \cos\beta \approx 1$, such that $v_B \approx v$. Due to Proposition 1, we can also show that $v_{ref} \cong v$ by examining (9), which shows that $v_{ref} = v_B \cos(\theta_e - \alpha_r) + y_e \dot{\theta}_{ref}$ to maintain $\dot{x}_e = 0$. Since $\theta_e, \alpha_r, y_e$, and $\dot{\theta}_{ref}$ are small, (9) reduces to $v_{ref} \cong v_B \cong v$.

## IV. KINEMATIC CONTROLLER

This section derives the kinematic controller. Sideslip compensation is presented first, followed by a path manifold based continuous VSC law providing graceful error convergence.

### A. Sideslip Compensation

Sideslip compensation is proposed to compensate for rear wheel sideslip, $\alpha_r$, in (12). Since an estimate of sideslip, $\hat{\beta}$, is produced by the observer where $\beta = K_F\hat{\beta}$ and $K_F = 1$ is the gain, and since the parameters in (20) are uncertain, we correlate $\alpha_r$ to $\hat{\beta}$ in (12) by defining,

$$\bar{\theta}_e = \theta_e + \beta = \theta_e + K_F\hat{\beta}. \tag{21}$$

$\bar{\theta}_e$ is used to design the controller. Proposition 3 and (21) are applied to (12) to produce,

$$\dot{y}_e = v\sin(\bar{\theta}_e - K_F\hat{\beta} - \alpha_r). \tag{22}$$

Since $\alpha_r$ and $\beta$ are typically different, the perturbation $\delta_{\alpha r} = \alpha_r + K_f\hat{\beta}$ describes the effect of the sideslip compensation:

$$\dot{y}_e = v\sin(\bar{\theta}_e - \delta_{\alpha r}). \tag{23}$$

which serves as the basis of the controller design. The effect of $\delta_{\alpha r}$ depends on vehicle speed. Since $\beta$ is required to analyze $\delta_{\alpha r}$, solving (19) for $\alpha_r$ results in $\alpha_r = \beta - L_r\kappa$. Substituting this into (20) and solving for $\beta$ results in,

$$\beta = \frac{\kappa}{C_r L}(C_r L L_r - mv^2 L_f) \tag{24}$$

Substituting (20) for $\alpha_r$ and (24) for $\beta$, into the above $\delta_{\alpha r}$ perturbation equation, results in,

$$\delta_{\alpha r} = \frac{\kappa}{C_r L}(C_r L L_r - 2mv^2 L_f). \tag{25}$$

Curvature, $\kappa$, is bounded by passenger comfort, which limits lateral acceleration to $0.32\, g$ (i.e., $3.13\, m/s^2$) [42] where centripetal acceleration is $a_y = v^2\kappa$. Since the maximum steering angle of the vehicle corresponds to $\kappa = 0.03\, m^{-1}$, the maximum curvature is $\kappa_{max} = \min(0.03, 3.13/v^2)$. As a result, curvature $\kappa$ decreases with speed, Fig 4. $\alpha_r$ saturates with increasing speed and $\beta$ approaches $\alpha_r$. Perturbation $\delta_{\alpha r}$ is initially a small positive number and becomes a small negative number as speed increases. Since the controller is designed based upon (23), $\delta_{\alpha r}$ causes the controller to slightly overcompensate for slip at low speeds to help reduce tracking error and slightly undercompensates at high speeds to make the controller more cautious. Given the estimated vehicle parameters in Sec VIII.C.3), slip compensation matches at ~9.5 m/s where $\delta_{\alpha r} = 0$.

### B. Kinematic Control Law

The kinematic controller aims to gracefully stabilize tracking error. Integrator state, $\sigma_k$, reduces tracking error:

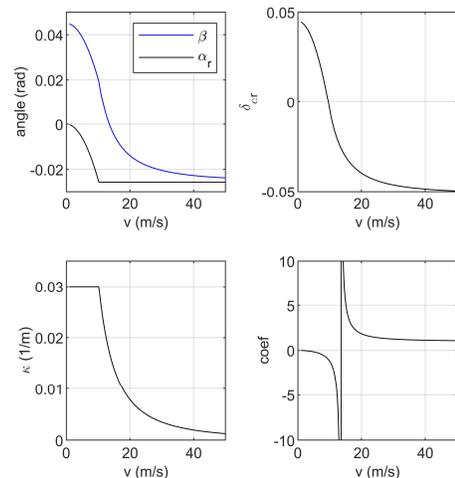

Fig 4: Maximum slip angles (top left) and perturbation, $\delta_{\alpha r}$, (top right) given vehicle speed for allowable curvature (bottom left) and coefficient relating $\alpha_r$ and $\beta$ per (20) (bottom right).

$$\dot{\sigma}_k = y_e. \quad (26)$$

A path manifold is designed based upon (23) and (26) to stabilize $y_e$ and $\sigma_k$ per VSC design, which is proposed in Proposition 4. A continuous controller is then designed in Theorem 1 to drive the system to the path manifold, eliminate chatter, and improve smoothness in practice.

**Proposition 4**: A path manifold function, $S_{kin}(t)$,

$$S_{kin}(t) = \bar{\theta}_e + \arcsin\left(\text{sat}\left(\left(\frac{c(t)y_e + K_i \sigma_k}{\bar{v}}\right), a_1\right)\right) \quad (27)$$

is proposed to guide error $x_1 = [\sigma_k \ y_e]^T$ asymptotically to $x_{1s} = [\sigma_{k.ss} \ 0]^T$. $c(t)$ is a slowly varying bounded function, $0 < c(t) \le c_{max}$, $K_i$ is the integral gain, and $a_1$ is the upper limit imposed by the saturation, sat(), to satisfy the limited range of the arcsine function, and $c_{max}$ is the upper limit on $c(t)$. This defines an unsaturated domain of the path manifold: $D_1 = \{x_1 \in \Re^2 | \left|\frac{c(t)y_e + K_i \sigma_k}{\bar{v}}\right| \le a_1 < 1 \}$. As in Sec III.B, $\bar{v} = \max(v_\epsilon, v_B)$, where $v_\epsilon$ is an arbitrary positive number to avoid singularity in (27) due to problematic zero velocity. ∎

Proof: We will now prove that (27) will drive the system (23) and (26) to an equilibrium point $x_{1ss} = [\sigma_{k,ss} \ y_{e,ss}]^T$ when the system is on the path manifold, $S_{kin} = 0$. First note that (23) can be transformed using trig identities to show that $\dot{y}_e = \bar{v} \sin(\bar{\theta}_e - \delta_{ar}) = \bar{v} \sin \bar{\theta}_e \cos \delta_{ar} - \bar{v} \cos \bar{\theta}_e \sin \delta_{ar}$. Since $\delta_{ar}$ and $\bar{\theta}_e$ are small, this can be simplified to $\dot{y}_e = v \sin \bar{\theta}_e - v \delta_{ar}$. Applying $\bar{v} = \max(v, v_\epsilon)$ for $\dot{y}_e$,

$$\dot{y}_e = \bar{v} \sin \bar{\theta}_e - \bar{v} \delta_{ar}. \quad (28)$$

On the manifold $S_{kin}(t) = 0$ and the state is inside the unsaturated domain $D_1$; thus $\bar{\theta}_e = -\arcsin\left(\frac{c(t)y_e + K_i \sigma_k}{\bar{v}}\right)$. Applying this to (23) and (26), the result is $\dot{y}_e = -c(t)y_e - K_i \sigma_k - \bar{v} \delta_{ar}$. Given (23) the equilibrium point is $y_{e,ss} = 0$ and $\sigma_{k,ss} = -\bar{v}\frac{\delta_{ar}}{K_i}$. If the vehicle parameters are constant, $\delta_{ar}$ will be a constant value. Otherwise, per (25), $\delta_{ar}$ will vary with changes in curvature, $\kappa$, velocity, $v$, and cornering stiffness, $C_r$.

Dynamics on the path manifold inside $D_1$ are then described by $\dot{x}_1 = \bar{A}x_1 + \begin{bmatrix} 0 \\ -\bar{v}\delta_{ar} \end{bmatrix}$, where $\bar{A} = \begin{bmatrix} 0 & 1 \\ -K_i & -c(t) \end{bmatrix}$ determines convergence to the equilibrium point. Since $K_i, c(t) > 0$, $\bar{A}$ is Hurwitz and the origin is exponentially stable. $\bar{v}\delta_{ar}$ is presumed to vary slowly. $c(t)$ varies slowly and is much greater than $K_i$. Thus, the path manifold stabilizes the majority of $y_e$ rapidly due to $c(t)$ and the integrator compensates for small remaining error as time progresses. While variations in $\bar{v}\delta_{ar}$ may cause $\sigma_{k,ss}$ to vary, the lateral error, $y_{e,ss}$, is still driven to zero, $y_{e,ss} \to 0$. For ideal conditions, if $v \cong 9.5$ m/s, then $\sigma_{k,ss} \to 0$.

If $\bar{\theta}_e$ is in a saturated domain, i.e., $\left|\frac{c(t)y_e + K_i \sigma_k}{\bar{v}}\right| > a_1$, the result is that $\bar{\theta}_e = -\arcsin(\pm a_1)$, where the sign of $a_1$ depends on whether the expression $\frac{c(t)y_e + K_i \sigma_k}{\bar{v}}$ is saturated from above or below. When applied to (27), $\dot{y}_e = \mp a_1 - \bar{v}\delta_{ar}$, which assures trajectories move toward and enter $D_1$ in finite time so long as $a_1$ is sufficiently large to dominate $v\delta_{ar}$: $|\bar{v}\delta_{ar}| < a_1 \le 1$. Given the range of $\delta_{ar}$, Fig 4, this suggests that vehicle states should not be in the saturated domain above 20 $m/s$, which is not a practical concern since initial conditions can be selected via path planning. The system enters $D_1$ in finite time whereupon exponential error convergence occurs inside $D_1$; the result is asymptotic convergence if the system starts in the saturated region. ∎

The continuous variable structure controller is now proposed.

**Theorem 1**: Defining $x_2 = [\bar{\theta}_e \ y_e \ \sigma_k]^T$, the kinematic yaw rate command, $r_{kin}$ for the VSC kinematic controller is:

$$r_{kin} = \kappa_{ref}\bar{v} + (\rho_{kin} + \psi_{kin}) \tanh\left(\frac{S_{kin}}{\varepsilon_{kin}}\right), \quad (29)$$

where $\psi_{kin} > 0$ is a small number to assure robustness, $\varepsilon_{kin}$ is a small number to control the boundary layer along the manifold (27), and

$$\rho_{kin} = \left|\frac{\dot{c}(t)y_e + c(t)\bar{v} \sin \bar{\theta}_e - c(t)\bar{v}\delta_{ar} + K_i y_e}{\bar{v}\sqrt{1-\left(\frac{c(t)y_e + K_i \sigma_k}{\bar{v}}\right)^2}}\right| \quad (30)$$

drives $S_{kin} \to 0$ in finite time and $x_2 \to [0, 0, \sigma_{k,ss}]$ asymptotically, which can further be proven to provide local exponential convergence of $x_2 \to [0, 0, \sigma_{k,ss}]$. ∎

Proof of Theorem 1 is in the appendix for brevity. Note that the transition created by $\tanh\left(\frac{S_{kin}}{\varepsilon_{kin}}\right)$ varies with different $\varepsilon_{kin}$: smaller $\varepsilon_{kin}$ reduces the unsaturated domain such that the shape of $\tanh\left(\frac{S_{kin}}{\varepsilon_{kin}}\right)$ approaches $\text{sign}\left(\frac{S_{kin}}{\varepsilon_{kin}}\right)$, which would lead to more aggressive convergence and chatter if $\varepsilon_{kin}$ is too small. Greater $\varepsilon_{kin}$ enlarges the unsaturated domain such that $\tanh\left(\frac{S_{kin}}{\varepsilon_{kin}}\right)$ is like a saturation function. As a result, the vehicle system response slows down to provide more graceful convergence. As highlighted in the proof of Theorem 1, tuning of $\psi_{kin}$ is related to the magnitude of $\varepsilon_{kin}$ to assure stability and convergence.

## V. TIME-VARYING PATH MANIFOLD

The path manifold parameter, $c(t)$, defined as the *convergence gain function*, determines how aggressively the vehicle follows the path. Design of $c(t)$ must consider vehicle safety and response. Path manifolds in [43] were geometrically based whereas this research considers vehicle speed, potential lateral error due to disturbances, and combined heading and lateral error.

### A. Dynamic Response Characteristics

Response characteristics, such as settling time or distance, are affected by $c(t)$ as described in Theorem 2 below.

**Theorem 2**: Settling time, $T_s$, and distance, $S_{path}$, can be related to $c(t)$ by $T_s \approx \frac{4}{c(t)}$ and $S_{path} = \bar{v}\frac{4}{c(t)}$.

Proof: Once the reaching phase has converged to the path manifold, $S_{kin}(t) = 0$. Per the state-space form of the system shown in the proof of Proposition 4, the poles of the system are $S_{1,2} = -\frac{c(t) \pm \sqrt{c(t)^2 - 4K_i}}{2}$. Since $c(t) \gg K_i$, they can be approximated as $S_{1,2} = -\epsilon, -c(t) + \epsilon$, where $\epsilon$ is a small positive number showing the effect of $c(t) \gg K_i$. Due to the small residual associated with $\epsilon$, the 98% settling time of the system is $T_s = \frac{4}{c(t)}$ where $c(t)$ is a slowly varying variable. This can also be used to assure convergence within a desired distance, $S_{path}$. If $\bar{v}$ varies relatively slowly, $S_{path} = \bar{v}T_s = \bar{v}\frac{4}{c(t)}$. ∎

Thus $c(t)$ determines convergence time and distance. Smaller $c(t)$ implies slower error convergence and provides more graceful motion in tracking error stabilization, which requires greater distances. $c(t)$ should decrease at higher speeds to allow more graceful convergence, which is described further in Sec V.D.

## B. Vehicle Safety Factors

For steering, vehicle safety relates to lateral stability and roll stability [43], and avoids lateral sliding or rollover due to excessive lateral acceleration. Lateral acceleration, $A = r_{kin}\bar{v} = \ddot{y}$ must be below a safety threshold, $A_{max}$, i.e., $|A| \leq |A_{max}|$. Lateral acceleration is limited by road-tire friction coefficient, $\mu$,

$$0 < |A_{max}| \leq k_1 \mu g, \quad (31)$$

where $k_1$ is a safety margin describing how much friction is allowed for steering, which can be adjusted to account for roll stability and braking. Given this limit, the upper bound on $c(t)$ can be established based upon tracking error, velocity, and system coefficients as described by Theorem 3.

**Theorem 3**: To satisfy vehicle safety associated with steering and road friction, $c(t)$ must be selected to satisfy,

$$c(t) \leq \frac{(k_1 k_2 \mu g - |\bar{v}\psi_{kin}|)\left(\sqrt{1-(a_1)^2}\right) - |K_i y_e| - |\dot{c}(t) y_e|}{|\bar{v}|(|\sin\bar{\theta}_e| + |\delta_{\alpha r}|)} \quad (32)$$

based upon anticipated operating conditions and typical tracking error. Coefficient $k_2$ balances control authority used for path characteristics and error correction where $0 < k_2 < 1$.

Proof: The magnitude of lateral acceleration is,

$$|\ddot{y}| = |\bar{v}r_{kin}| = \left|\kappa_{ref}\bar{v}^2 + \bar{v}(\rho_{kin} + \psi_{kin})\tanh\left(\frac{S_{kin}}{\varepsilon_{kin}}\right)\right| \quad (33)$$

where (29) is substituted for $r_{kin}$. The right hand side can be expanded to: $|\ddot{y}| \leq |\kappa_{ref}\bar{v}^2| + |((\rho_{kin} + \psi_{kin})\tanh(\frac{S_{kin}}{\varepsilon_{kin}}))\bar{v}| \leq |A_{max}|$. The $|\kappa_{ref}\bar{v}^2|$ part is based upon the path curvature $\kappa_{ref}$ where a portion of $A_{max}$ specified by $k_2$ is reserved as follows,

$$|\kappa_{ref}\bar{v}^2| < (1 - k_2)A_{max} \quad (34)$$

Subtracting this from (33), lateral error acceleration is then,

$$|\ddot{y}_e| \approx \left|((\rho_{kin} + \psi_{kin})\tanh(\frac{S_{kin}}{\varepsilon_{kin}}))\bar{v}\right| \leq k_2 |A_{max}|. \quad (35)$$

Eliminate $\rho_{kin}$ using (30) given the worst case $\left|\tanh\left(\frac{S_{kin}}{\varepsilon_{kin}}\right)\right| = 1$ and $\arcsin(\cdot)$ replaced by $a_1$ per (27), which provides,

$$\rho_{kin} \leq \frac{|\dot{c}(t)y_e| + |c(t)\bar{v}\sin\bar{\theta}_e| + |c(t)\bar{v}\delta_{\alpha r}| + |K_i y_e|}{\bar{v}\sqrt{1-a_1^2}}, \quad (36)$$

Substituting (36) into (35), noting that $|\tanh()| \leq 1$ and $|a+b| < |a| + |b|$, the result is that,

$$\ddot{y}_e \leq \frac{|\dot{c}(t)y_e| + |c(t)\bar{v}\sin\bar{\theta}_e| + |c(t)\bar{v}\delta_{\alpha r}| + |K_i y_e|}{\sqrt{1-a_1^2}} + |\bar{v}\psi_{kin}| < k_2 A_{max}, \quad (37)$$

must be true. Using the upper limit for $A_{max}$ in (31) and solving for $c(t)$, the upper limit of $c(t)$ in (32) results. ∎

Theorem 3 highlights that the upper limit of $c(t)$ depends on heading error and lateral error, as well as velocity. While a lateral perturbation due to wind or path discontinuity may cause $y_e$ to increase initially, $\theta_e$ will vary as the vehicle steers back to the desired path, but the magnitude of $\theta_e$ variations depend on the magnitude of $c(t)$. Simulation results in Sec V.D highlight appropriate $c(t)$ for different velocities and lateral perturbations.

## C. Steering Actuator Dynamic Response

Steering actuator rate saturation is important to consider when determining $c(t)$. Convergence within a desired time or distance requires more aggressive steering at low speeds, whereas higher speeds require less aggressive steering; both are governed by $c(t)$.

**Proposition 5**: Larger ranges of posture error, $y_e$ and $\theta_e$, can be accommodated using smaller $c(t)$ at low speeds, whereas higher speeds up to 10 m/s allow larger $c(t)$ without steering saturation. ∎

Proof: From (16), note that yaw rate can be expressed as $r_{KIN} = \frac{v_B}{L}\tan\varphi$ if nonslip is assumed. Solving for steering angle, $\varphi$, and taking the time derivative, $\omega = \dot{\varphi} = \frac{\dot{r}_{KIN}\bar{v}L}{\bar{v}^2 + r_{KIN}^2 L^2}$. Thus, $\omega$ is dependent upon yaw rate and acceleration, $r_{KIN}$ and $\dot{r}_{KIN}$, commanded by (27), (30), and (29). This results in a complex expression [39] highlighting that steering rate depends on $y_e$, $\bar{\theta}_e$, $\bar{v}$, $c(t)$, and $\sigma_k$, which is studied numerically here.

Fig 5 presents contour plots as functions of $y_e$ and $\theta_e$ where regions with unsaturated steering rate are shaded based upon maximum steering rate of $\pm 0.3 \, rad/sec$, $\varepsilon_{kin} = 0.1$, $\psi_{kin} = 0.1$, $\sigma_k = 0$, and $\kappa_{ref} = 0$ typical for converging to a straight path. At lower velocities, smaller $c(t)$ allows for a larger set of initial errors without steering saturation, Fig 5 $(a) - (f)$. The unsaturated region increases with higher velocity for a given value of $c(t)$, Fig 5 $(g) - (i)$. This leads to the general conclusion that lower velocities are better served by smaller $c(t)$, while larger velocities up to 10 m/s allow larger $c(t)$. ∎

## D. Selecting and Varying $c(t)$

Selecting $c(t)$ depends on the speed of the vehicle, potential posture error encountered during operation, and desired gracefulness.

We first simulate the system at a constant speed with different lateral perturbations, $y_e$, to find constant values of $c(t)$ that allow the vehicle to gracefully stabilize back to a straight path. Fig 6 highlights lateral error convergence and acceleration for a speed of $10 \, m/s$. Fig 6 (row 1) highlights that larger $c(t)$ is more aggressive, converging more quickly with larger acceleration, whereas smaller $c(t)$ is less aggressive and more graceful. Fig 6 (row 2) simulates the vehicle perturbed laterally at $10 \, m/s$ with constant $c(t) = 2$, highlighting that larger perturbations result in more aggressive compensation and lateral acceleration.

These results are extended in Fig 7 for speeds varying between 1 to 40 $m/s$ to show $c(t)$ values producing similar gracefulness as shown in Fig 6 (row 1) for different lateral perturbations. Larger values of $c(t)$ are appropriate at lower speeds (e.g, $5 - 10 \, m/s$)

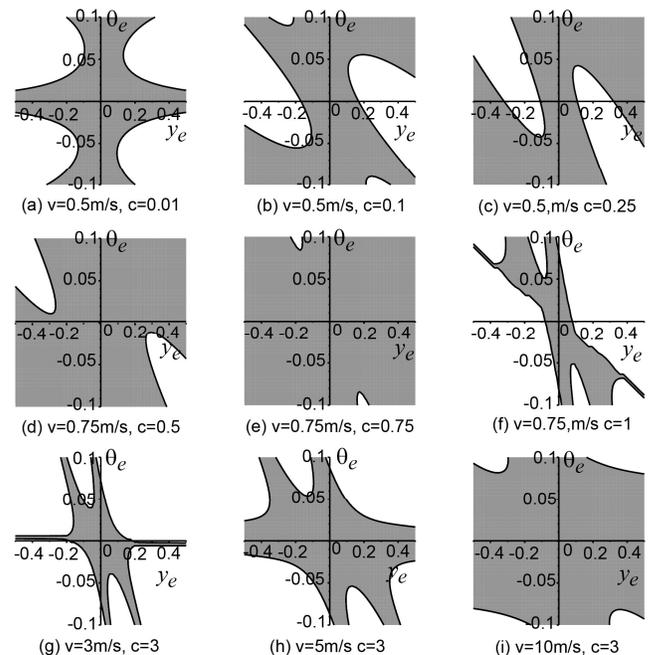

(a) v=0.5m/s, c=0.01  (b) v=0.5m/s, c=0.1  (c) v=0.5,m/s c=0.25
(d) v=0.75m/s, c=0.5  (e) v=0.75m/s, c=0.75  (f) v=0.75,m/s c=1
(g) v=3m/s, c=3  (h) v=5m/s, c=3  (i) v=10m/s, c=3

Fig 5: Steering rate saturation. Shaded regions highlight where steering rate will not be saturated.

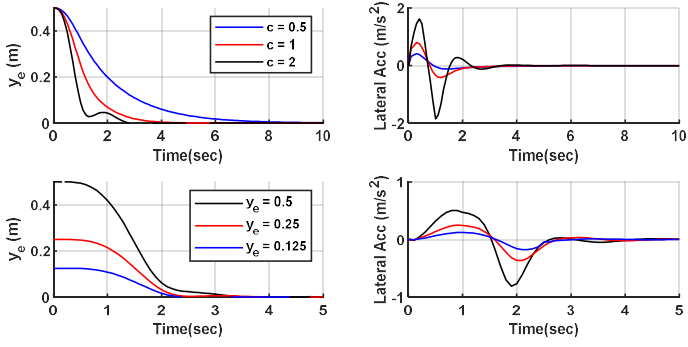

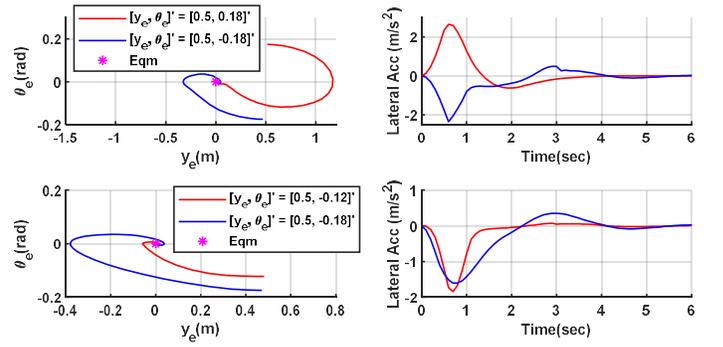

Fig 6. The first row uses different constant $c(t)$ to highlight its effect on convergence at $10\ m/s$ whereas the second row highlights the effect of perturbation magnitude $y_e$ on convergence at $10\ m/s$ with $c = 2$.

Fig 8: Tracking error phase portraits (left) and lateral acceleration plots (right) using fixed values for $c_0$ (1st row) and $c_0$ based upon intial conditions (2nd row) and zero initial velocity. Both rows use $t_{end} = 4$ s.

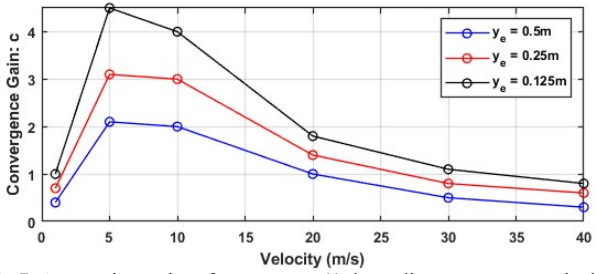

Fig 7. Appropriate values for constant c(t) depending on constant velocity magnitude and potential lateral perturbations, $y_e$.

and with smaller lateral error perturbations, which supports Theorem 2 and Theorem 3. $c(t)$ should be smaller with much smaller speeds (e.g. $< 5\ m/s$) or higher speeds ($> 10\ m/s$) or larger perturbations. At 10 m/s, a constant $c(t) = 3$ is appropriate for perturbations of $y_e(0) = 0.25$ m, which is studied extensively in the physical tests. Transitioning between values of $c(t)$ is now presented in following proposition.

**Proposition 6**: *Hierarchal Path Manifolds:* It is proposed to vary $c(t)$ linearly with time from an initial value, $c_0$, to a final value, $c_{ss}$:

$$c(t) = c_{ss}\frac{t}{t_{end}} + c_0\left(1 - \frac{t}{t_{end}}\right), \quad (38)$$

where $c(t_0) = c_0$ and $c(t_{end}) = c_{ss}$, where $t_0 \leq t \leq t_{end}$, $t_{end} = t_0 + \Delta t$. $\Delta t$ is the desired transition time from $c_0$ to $c_{ss}$. To satisfy safety constraints, $c_0$, $c_{ss}$, and $c(t)$ must be positive and satisfy (32). It is proposed that $c(t) = \min((38), (32))$. ∎

Simulations illustrate the effect of varying $c(t)$ where the vehicle starts with an initial posture error and zero velocity, then accelerates to driving speed while converging to a straight path. The final speed of the vehicle is $10\ m/s$, so $c_{ss} = 3$ is used per Fig 7. A transition time of $t_{end} = 4$s is applied. Two techniques were examined for selecting $c_0$: 1) a small fixed $c_0$ and 2) a computed $c_0$ based on (27) to make $S_{kin} = 0$ at the initial posture.

In 1), a small fixed $c_0$ is applied since velocity starts at zero and error is large initially. This allows the system to gracefully converge as velocity increases, Fig 8 (row 1), where a phase portrait of lateral error and heading error is shown on the left and lateral acceleration on the right. Two initial postures are applied. The red curve shows $y_e$ and $\bar{\theta}_e$ with the same sign, while the signs are different in the blue curve. Lateral error, $y_e$, increases for the red curve since the vehicle is initially pointed away from the path. Thus, it has more aggressive motion than the blue one since the vehicle is initially pointed towards the path in that case. This occurs if initial posture error is in Quadrant I and III of the phase portraits. Selecting $c_0$ to be small initially assures that the path manifold (27) provides smooth and graceful convergence.

In the case 2), we use (27) to find $c_0$ to align the path manifold with the initial posture, Fig 8 (row 2). Comparing the blue trajectories in the first and second rows, the trajectory using this technique has $c_0 = 0.036$ and results in smaller acceleration and smoother convergence since 2) eliminates convergence to the manifold. Heading error is reduced in the red trajectory, which results in $c_0 = 0.024$, but results in more rapid convergence. Compared to 1) (first row), $c_0$ based on (27) in 2) provides more graceful motion than the manually selected one. However, the application of (27) in computing $c_0$ requires initial posture in Quadrant II and IV, or on the $y_e$ axis. Otherwise, manual selection of $c_0$ per 1) must be used.

It is recommended to use Fig 7 to determine appropriate $c(t)$ during operation based upon perturbations and operating speeds. Proposition **6** provides a guide for varying $c(t)$ during vehicle startup and driving at varying speeds.

## VI. DYNAMIC CONTROLLER

The dynamic controller is low-level in the multi-tiered structure. It steers the vehicle to create the desired yaw rate specified by the high-level kinematic controller. The dynamic controller contains a two-tiered closed-loop controller to stabilize yaw tracking error and then steering tracking error using backstepping.

### A. Yaw-tracking Control

The yaw-tracking controller is the first level of the dynamic controller with the goal of finding a desired steering angle, $\varphi_{des}$. The reference yaw rate $r_{kin}$ is provided by (29), while the yaw rate from the dynamics is $r_{dyn}$. The yaw tracking error is defined as $r_e = r_{kin} - r_{dyn}$. Taking the time derivative and substituting (15) for $\dot{r}$ results in the yaw error dynamics,

$$\dot{r}_e = -a_{21}\beta - a_{22}r_e + (\dot{r}_{kin} - a_{22}r_{kin}) - b_{21}\varphi_{des} - \delta_r. \quad (39)$$

where coefficients $a_{21}$, $a_{22}$, and $b_{21}$ are from Proposition 2. A steering controller is now proposed to provide convergence.

**Theorem 4**: Exponential convergence of yaw rate error dynamics (39) is provided by the steering angle command,

$$\varphi_{des} = -\frac{a_{21}\beta - \dot{r}_{kin} + a_{22}r_{kin} - K_{p1}r_e - K_{i1}\sigma_r}{b_{21}} \quad (40)$$

where $\sigma_r$ is the integral of yaw rate error, $K_{p1} > a_{22}$, and $K_{i1} > 0$.

Proof: The integral of yaw rate error is provided by $\dot{\sigma}_r = r_e$. Substituting (40) into (39) results in the yaw error dynamics,

$$\dot{r}_e = -(K_{p1} + a_{22})r_e - K_{i1}\sigma_r + \delta_r. \quad (41)$$

Combined with the integrator state equation, the system can be expressed in matrix form using the state vector $x_3 = [\sigma_r \ r_e]^T$ [39]. Given $\delta_r$ varies slowly, Theorem 4 assures that the system is Hurwitz and $x_3$ is exponentially stable at $x_{3,ss} = [\sigma_{r,ss} \ 0]^T$. ∎

### B. Steering-tracking Control

The steering actuator controller must be designed with the dynamic controller due to the close coupling between steering actuator response and vehicle dynamics. Steering-tracking control provides yaw-tracking control using steering rate, $\omega$. The difference between the desired and actual steering angle is defined as the *steering angle error*, $\varphi_e = \varphi_{des} - \varphi_{act}$. **Backstepping control** leverages dynamic extension such that steering rate, $\omega$, becomes the dynamic control input for improved tracking instead of traditional steering angle in [8, 35]. The resulting backstepping controller is derived in the following theorem.

**Theorem 5**: Backstepping steering tracking control provides asymptotic convergence of yaw rate error, $r_e$, and steering error, $\varphi_e$, using,

$$\omega = -\frac{a_{21}\beta + (a_{22}\dot{r}_{kin} - \ddot{r}_{kin}) - K_{p1}\dot{r}_e - (K_{i1} + b_{21})r_e}{b_{21}} + K_{p2}\varphi_e + K_{i2}\sigma_\varphi \quad (42)$$

where $\sigma_\varphi$ is the integral of $\varphi_e$, $K_{p2} > a_{22}$, and $K_{i2} > 0$. ∎

The proof is in the appendix for the interested reader.

## VII. OUTPUT FEEDBACK CONTROL

Since sideslip cannot be measured directly, output feedback control is achieved using an observer and replacing the states in the controllers with observer-based estimates. The dynamic model can be represented in the classical triangular normal form [38, 44] and is locally Lipschitz, so a High-Gain Observer (HGO) is applied to provide robustness. Defining the dynamic state $\underline{x} = [r_{dyn} \ \beta]^T$, the dynamic model (15) and (14) is represented by $\dot{r}_{dyn} = \phi_r(\underline{x}, \varphi) = \phi_{r0}(\underline{x}, \varphi) + \beta + \delta_r'$ and $\dot{\beta} = \phi_\beta(\underline{x}, \varphi) = \phi_{\beta_0}(\underline{x}, \varphi) + \delta_\beta$ where $\delta_r' = \delta_r - \beta$. $\phi_{\beta_0}(\underline{x}, \varphi)$ and $\phi_{r0}(\underline{x}, \varphi)$ are the linear terms from (14) and (15) that form the *linearized dynamic model*.

The measured output is $y = [1 \ 0] \underline{x} = r_{dyn}$. The system can be shown to be observable since $a_{22} \neq 0$. The locally Lipschitz state feedback control law (40) that stabilizes $\underline{x}$ is denoted by $\varphi(\underline{x})$. Applying $\varphi(\underline{x})$ to the linearized dynamic model results in $\dot{\beta} = \phi_{\beta_0}(\underline{x}, \varphi(\underline{x}))$ and $\dot{r}_{dyn} = \phi_{r0}(\underline{x}, \varphi(\underline{x}))$. The observer estimates the state $\hat{\underline{x}} = [\hat{r}_{dyn} \ \hat{\beta}]^T$ as follows:

$$\dot{\hat{r}}_{dyn} = \phi_{r0}(\hat{\underline{x}}, \varphi(\underline{x})) + h_1(y - \hat{r}_{dyn}) \quad (43)$$
$$\dot{\hat{\beta}} = \phi_{\beta_0}(\hat{\underline{x}}, \varphi(\underline{x})) + h_2(y - \hat{r}_{dyn}) \quad (44)$$

where $h_1$ and $h_2$ are the observer gains. Estimation error is $\tilde{\underline{x}} = [\tilde{r}_{dyn} \ \tilde{\beta}]^T$, where $\tilde{r} = r_{dyn} - \hat{r}_{dyn}$ and $\tilde{\beta} = \beta - \hat{\beta}$. Substituting (14), (15), (43) and (44) results in the estimation error model,

$$\dot{\tilde{r}}_{dyn} = -h_1\tilde{r}_{dyn} + \tilde{\beta} + \delta_r'(\underline{x}, \tilde{\underline{x}}) \quad (45)$$
$$\dot{\tilde{\beta}} = -h_2\tilde{\beta} + \delta_\beta(\underline{x}, \tilde{\underline{x}}) \quad (46)$$

The scaled estimation errors are defined as $\eta_1 = \tilde{r}_{dyn}/\varepsilon$ and $\eta_2 = \tilde{\beta}$, where $\varepsilon$ is a small positive number tuned to optimize observer convergence. Substituting these terms into the estimation error model, the resulting *scaled estimation model* is then,

$$\varepsilon\dot{\eta}_1 = -\alpha_1\eta_1 + \eta_2 + \delta_r'(\underline{x}, \tilde{\underline{x}}) \quad (47)$$
$$\varepsilon\dot{\eta}_2 = -\alpha_2\eta_1 + \varepsilon\delta_\beta(\underline{x}, \tilde{\underline{x}}) \quad (48)$$

where $\alpha_1$ and $\alpha_2$ are positive observer gains such that $h_1 = \alpha_1/\varepsilon$ and $h_2 = \alpha_2/\varepsilon^2$. The characteristic equation of the scaled estimation model is then,

$$s^2 + \alpha_1 s + \alpha_2 = 0 \quad (49)$$

such that $\alpha_1$ and $\alpha_2$ are selected for desired response characteristics. Thus, (47) and (48) suggest that the impact of $\delta_\beta(\underline{x}, \tilde{\underline{x}})$ and $\delta_r(\underline{x}, \tilde{\underline{x}})$ are greatly reduced for small positive $\varepsilon \ll 1$. Convergence of scaled estimation error $\eta_1$ and $\eta_2$ are also much faster than the regular error terms $\tilde{\beta}$ and $\tilde{r}$ for $\varepsilon \ll 1$.

The output feedback dynamic controller is then,

$$\hat{r}_e = r_{kin} - \hat{r}_{dyn} \quad (50)$$
$$\hat{\varphi}_{des} = -\frac{a_{21}\hat{\beta} - \dot{r}_{kin} + a_{22}r_{kin} - K_{p1}\hat{r}_e - K_{i1}\hat{\sigma}_r}{b_{21}} \quad (51)$$
$$\hat{\varphi}_e = \hat{\varphi}_{des} - \varphi_{act} \quad (52)$$
$$\dot{\hat{r}}_e = -(K_{p1} + a_{22})\hat{r}_e - K_{i1}\hat{\sigma}_r \hat{r}_e + \hat{\varphi}_e \quad (53)$$
$$\hat{\omega} = -\frac{a_{21}\dot{\hat{\beta}} - \ddot{r}_{kin} + a_{22}\dot{r}_{kin} - K_{p1}\dot{\hat{r}}_e - (K_{i1} + b_{21})\hat{r}_e}{b_{21}} + K_{p2}\hat{\varphi}_e + K_{i2}\hat{\sigma}_\varphi \quad (54)$$

where (54) is the output feedback control law for the dynamic steering controller. Derivatives of $r_{kin}$ are calculated analytically based upon (29). Derivatives of $\hat{r}_e$ are based upon (39). Stability of the overall system is guaranteed by the separation principle [45], which indicates that small positive $\varepsilon \ll 1$ allows the observer and controller to be designed independently since the observer rejects perturbations from the controller and uncertainty. Peaking is resolved as described in the next section.

## VIII. EVALUATION PROCEDURES

Field experiments are used to evaluate the research. The platform, test fields and paths, metrics, and baseline controllers are presented.

### A. Vehicle Platform

The platform is Red Rover; a 2005 Dodge Caravan, Fig 9, with a modified Kairos ProntoIV autonomy system to accept both steering angles and rates. A NovaTel GPS with Omnistar HP differential service senses posture with $\sim 0.1 \ m$ and $0.2 \ deg$ accuracy. A digital gyroscope was added to sense yaw rate. Steering wheel angle is measured by a steering servo with a range of $+/- 550º$, which correlates to a steering angle of $+/- 35º$. Vehicle speed is controlled by a human driver.

### B. Test fields and Paths

Two test fields in Salt Lake City, Utah are used, Fig 9: State Fair Park (SFP) and Merrell Engineering Building (MEB) parking. MEB provides a 10% sloped asphalt surface providing a gravity disturbance, whereas SFP provides a non-sloped uneven/bumpy gravel surface. MEB allows middle speeds ($\sim 5 - 7 \ m/s$) and SFP allows higher speeds ($\sim 10 \ m/s$).

Two basic paths were evaluated in both fields. A Reed-Shepps "L" shaped path demonstrates response to straight paths, constant curvature paths, and curvature discontinuity. A $40 \ m$ line is followed by a $50 \ m$ radius arc sweeping out $90°$ and is concluded by another $40 \ m$ line. A graceful 'S' shaped Euler spiral with continuously varying curvature demonstrates realistic path variation. Radius varies from a $100 \ m$ radius right to left.

The SFP comprehensive path, Fig 9 and Table 1, examines more complex paths at higher speeds. A bumpy gravel surface is used. Path segments of sufficient length assure convergence.

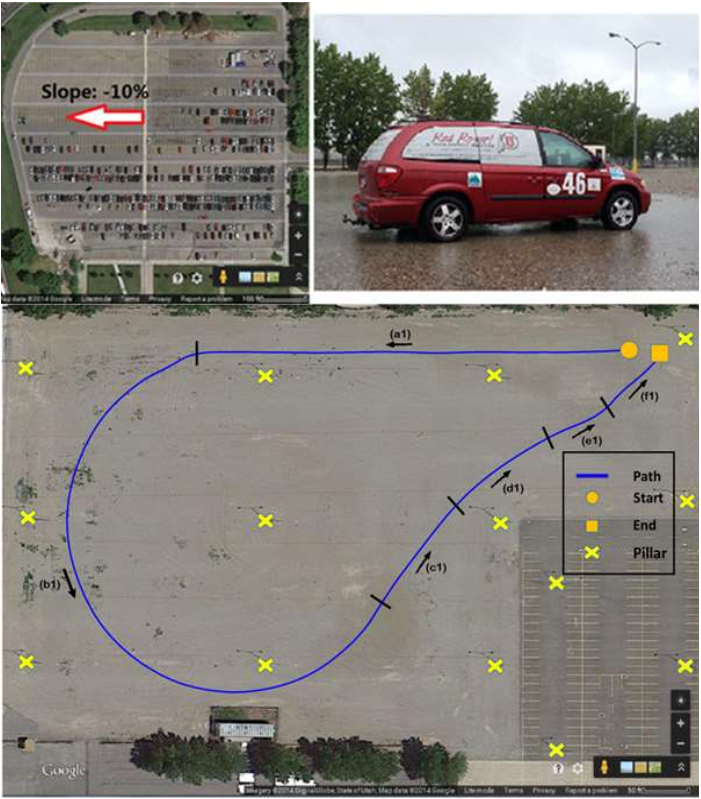

Fig 9: Experimental Test Fields: Merrill Engineering Building (MEB) parking lot (Upper Left), Utah State Fair Park (SFP) (Bottom), and Red Rover at SFP on a rainy day.

Table 1: Comprehensive path characteristics.

| SEG | Path Type | Path Settings | Length |
|---|---|---|---|
| a1 | Straight Line | L = 120 m | 120 m |
| b1 | Arc Curve | R = 50m, angle = 225° | ≈ 196.4 m |
| c1 | Euler's Spiral | κ = 0.02 → 0, angel = 10° | ≈ 17.5 m |
| d1 | Euler's Spiral | κ = 0 → -0.01, angel = 10° | ≈ 34.9 m |
| e1 | Arc Curve | R = -100m, angle = 20° | ≈ 17.5 m |
| f1 | Arc Curve | R = 100m, angle = 20° | ≈ 17.5 m |

### C. Controllers

Baseline controllers highlight features of the proposed work.

#### 1) Baseline A: Multi-tiered PID Steering Controller

Kinematics-based PID control [11, 12, 22] was explored, but performed poorly. *Multi-tiered PID control in the kinematic and dynamic tiers is proposed*. PID gains are tuned via traditional empirical tuning [46] to achieve a 4 sec settling time, $T_s$, with critical damping. Kinematic gains are tuned first, followed by dynamics designed to be twice as fast. The dynamic controller was tuned using a constant 0.1 rad/sec, yaw rate command at 10 m/s, e.g., 0.01 $m^{-1}$ path curvature. Specific gains are not listed due to vehicle specific variability.

#### 2) Baseline B: Robust Steering Controller

"B" is from [10], which is the basis of this work, but without slip-based kinematics and peaking compensation:

$$S_{kin} = \arcsin\left(\frac{c(t)y_e + K_i\sigma_k}{\bar{v}}\right) + \theta_e \quad (55)$$

$$\rho_{kin} = \left| -\kappa v + \frac{\dot{c}(t)\frac{y_e}{\bar{v}} + c(t)\sin\theta_e + \frac{K_i y_e}{\bar{v}}}{\sqrt{1-\left(\frac{c(t)y_e + K_i\sigma_k}{\bar{v}}\right)^2}} \right| \quad (56)$$

$$r_{kin} = -(\rho_{kin} + \psi_{kin})\tanh\left(\frac{S_{kin}}{\varepsilon_{kin}}\right) \quad (57)$$

where the output feedback dynamic control law is,

$$\hat{r}_e = \hat{r} - r_{ref} \quad (58)$$

$$\hat{\varphi}_e = \varphi_{act} - \hat{\varphi}_{des} = \varphi_{act} + \frac{a_{21}\hat{\beta} + a_{22}r_{ref} - \dot{r}_{ref} + K_p\hat{r}_e}{b_{21}} \quad (59)$$

$$\hat{\omega} = -\frac{a_{21}\hat{\beta} + a_{22}\dot{r}_{ref} - \ddot{r}_{ref} + K_{p1}\dot{\hat{r}}_e + b_{21}\hat{r}_e}{b_{21}} - K_{p2}\hat{\varphi}_e. \quad (60)$$

Tuning is similar to the proposed controller in the next section.

#### 3) Proposed Controller

The kinematic controller law is (23), (26), (27), (28), (29), and (30), and the output state feedback dynamic control is provide by the integrators and (50), (52), and (54). *Saturation is applied in the kinematic controller to limit peaking and yaw rate commands*:

$$r_{kin} = \text{sat}\left(\kappa_{ref}v_{ref} + (\rho_{kin} + \psi_{kin})\tanh\left(\frac{S_{kin}}{\varepsilon_{kin}}\right), r_{threshold}\right). \quad (61)$$

$r_{threshold} = 0.3\ rad/sec$ was calculated based upon allowable lateral acceleration given the current vehicle operating speed.

The dynamic controller was tuned first, followed by the kinematic controller. The steering angle tracking controller was designed to be twice as fast as the desired dynamic controller, which was twice as fast as the kinematic controller. The steering angle gains $K_{p2}$ and $K_{i2}$ were designed for a 1 sec setting time. The yaw rate tracking controller $K_{p1}$ and $K_{i1}$ were designed to satisfy desired dynamic controller performance. Nominal parameters of $C_f = 230KN/rad$, $C_r = 200KN/rad$, $m = 2540kg$, $J = 5000\ kg$–$m^2$, $L_f = L_r = 1.5m$, and $\mu = 0.8$ were used for design while perturbed parameters $C_f = C_r = 110KN/rad$, $m = 2300kg$, $J = 4500\ kg - m^2$, $L_f = 1.4m$, $L_r = 1.6m$ and $\mu = 0.8$ were used in simulations for evaluating performance.

Kinematic controller tuning requires selection of $c(t)$, $\psi_{kin}$, $\varepsilon_{kin}$, $K_i$, and $a_1$, which are dependent upon the safety factors $k_1$ and $k_2$, vehicle speed $\bar{v}$, and coefficient of friction, $\mu$. Worst case conditions are considered for calculating safety margins (e.g. $\bar{v} = 10\ m/s$ and $\mu = 0.5$) [40]. Assuming $k_1 = 0.8$ allowing for speed variations, (34) can be used to show $k_2 < 0.49$ for wet conditions, but $k_2 < 0.64$ for dry conditions (e.g. $\mu = 0.7$).

Selection of $c(t)$ is based upon (32) in Theorem 3. An initial estimate for $c(t)$ with 0.5 m initial lateral error and the parameters above is $c(t) < 3.4$ for rainy conditions. Using $c(t) = 3$ for tuning, simulations [39] and experiments indicated that $\psi_{kin} = 0.1$ and $\varepsilon_{kin} = 0.1$ balanced convergence and robustness. Larger $\psi_{kin}$ improves robustness whereas smaller $\varepsilon_{kin}$ makes the system more aggressive and more prone to oscillation and chatter. $K_i = 0.1$, which is much less than $c(t)$, provides gradual final error convergence. Sliding surface saturation $a_1 = 0.9$ avoids singularities while allowing more aggressive compensation.

Proposition 6 was used to vary $c(t)$ to deal with low initial speeds. $c(t)$ was based upon a combination of $c_0$ and $c_{ss} = 3.0$.

### D. Experimental Procedures

Field experiments evaluate the controllers using the paths, surface materials, and speeds described in Sec. VIII B. Initial posture error was $[\bar{\theta}_e\ \ y_e]^T = [0\ \ 0.5]^T$ with the vehicle starting from rest. Each controller and path were evaluated ten times. Basic paths are evaluated using both dry and wet surfaces (e.g., clear and rainy days) using lower speeds. The higher speed SFP comprehensive path was used with dry and wet gravel.

### E. Performance Metrics

Metrics characterize performance along each path segment (SEG). RMS lateral error, $E_{RMS}$, evaluates overall accuracy. Range of lateral error, $E_{RNG}$, and RMS error from the last 10 data points of a path segment, $E_{L1}$, characterize maximum error variation and final error for each SEG. Convergence success rate, %$C$, indicates the percent of trials converging lateral error to within 0.1m along a path segment. Higher %$C$ indicates that a controller can reliably converge, which is challenging in some of the shorter path segments indicated in Table 1. If %$C$ is zero, lateral error did not converge sufficiently before the SEG end. RMS lateral acceleration, $A_{RMS}$, relative to the reference lateral acceleration indicates jerkiness and gracefulness, denoted as Graceful Motion (G.M). Lateral acceleration is calculated from accelerometer data with moving window spline segments [47]. Averages ("$avg$") and standard deviation ("$\sigma$") are reported.

## IX. RESULTS

Results from the proposed and baseline controllers are presented where Baseline A, Baseline B, the proposed controller, and the proposed controller with saturation are referred to as "A", "B", "PROP", and "PROP-S".

### A. SFP Comprehensive Path Results

Data is provided for higher (8~10 $m/s$) and middle speeds (5.5~7.5 $m/s$), Table 2. Higher speed "A" results are not included due to unstable oscillations.

At higher speeds, performance of PROP and PROP-S is appreciated compared to Baseline B. B failed to converge lateral error on SEG $a_1$ in time and %$C = 0$, but it could proceed on the path. $E_{L10}$ for PROP and PROP-S were each 78% smaller than B. PROP-S has smaller $E_{RNG}$ and $E_{RMS}$ than PROP or B, which indicates PROP-S provides reduced overshoot and oscillation. PROP and PROP-S had ~62% and ~78% less $E_{L1}$ on $e_1$ and had ~29% and ~43% less $E_{L1}$ on $f_1$, respectively. B failed to converge lateral error on the continuously varying curvature SEGs, $c_1$ and $d_1$, and had larger lateral acceleration, $A_{RMS}$, than PROP and PROP-S. *It is easily appreciated that at high speeds PROP-S provides the best performance*, followed by PROP, and in a distant third place Baseline B. Results from middle speeds are improved for all controllers. PROP-S performs best, followed by PROP, B, and A.

The importance of the peaking saturation can be appreciated by comparing PROP and PROP-S high speed results, Table 2. Initial error converges to near zero on the straight SEG $a_1$ for both controllers (e.g. $E_{L10} \leq 0.1m$), but PROP only converges 60% of the time (%$C = 60\%$) before the SEG ends. PROP-S converges every time (%$C = 100\%$) since the yaw rate saturation in the controller limits peaking. As a result, PROP-S has 33% less $E_{L10}$ on $a_1$ compared to PROP. *Discontinuity* at the beginning of $b_1$ causes oscillations in $E_{ss}$, steering commands, and lateral acceleration, Fig 10 and Fig 11, but PROP-S has better transient response. PROP and PROP-S are perturbed, but both provide high tracking accuracy ($E_{L10} \approx 0.1m$) and always converge. Due to the yaw rate saturation, Fig 11, PROP-S has smaller $E_{RNG}$ (~11% less), $E_{RMS}$ (~16% less), and $A_{RMS}$ (~15% less) compared to PROP. Standard deviation of PROP-S $E_{L10}$ is 50% smaller than PROP. Thus, **PROP-S produces more accurate, smoother, and more consistent results, especially with path discontinuities.**

Table 2: Experimental results: steering performance comparison for all three steering controllers in SFP – Parking Lot at middle and higher speeds on the comprehensive path.

| Seg. (m) | Comprehensive Path | | High Speed (8~10 m/s) | | | Middle Speed (5.5~7.5 m/s) | | | |
|---|---|---|---|---|---|---|---|---|---|
| | | | B | PROP | PROP-S | A | B | PROP | PROP-S |
| | | | avg ± σ | avg ± σ | avg ± σ | avg ± σ | avg ± σ | avg ± σ | avg ± σ |
| a1 | Lateral Error (m) | $E_{RNG}$ | 0.79±0.11 | 0.73±0.09 | 0.72±0.08 | 0.88±0.08 | 0.70±0.21 | 0.74±0.08 | 0.73±0.07 |
| | | $E_{RMS}$ | 0.22±0.06 | 0.20±0.05 | 0.19±0.02 | 0.14±0.02 | 0.19±0.04 | 0.19±0.08 | 0.20±0.06 |
| | | $E_{L10}$ | 0.09±0.02 | 0.10±0.06 | 0.07±0.04 | 0.08±0.02 | 0.06±0.03 | 0.06±0.02 | 0.03±0.05 |
| | | %C | 0% | 60% | 100% | 90% | 100% | 100% | 100% |
| | G.M. (m/s^2) | $A_{RMS}$ | 0.24±0.10 | 0.21±0.06 | 0.21±0.07 | 0.21±0.02 | 0.17±0.08 | 0.11±0.03 | 0.11±0.03 |
| b1 | Lateral Error (m) | $E_{RNG}$ | 1.21±0.38 | 1.31±0.39 | 1.10±0.3 | 0.69±0.08 | 0.36±0.06 | 0.37±0.05 | 0.36±0.07 |
| | | $E_{RMS}$ | 0.44±0.04 | 0.32±0.06 | 0.27±0.07 | 0.17±0.01 | 0.20±0.02 | 0.05±0.01 | 0.06±0.02 |
| | | $E_{L10}$ | 0.46±0.08 | 0.1±0.04 | 0.10±0.02 | 0.09±0.07 | 0.17±0.03 | 0.06±0.02 | 0.05±0.02 |
| | | %C | 80% | 100% | 100% | 60% | 100% | 100% | 100% |
| | G.M. (m/s^2) | $A_{RMS}$ | 0.77±0.32 | 0.75±0.17 | 0.64±0.15 | 0.33±0.12 | 0.29±0.06 | 0.19±0.02 | 0.19±0.02 |
| c1 | Lateral Error (m) | $E_{RNG}$ | 0.71±0.11 | 0.49±0.07 | 0.43±0.15 | 0.33±0.12 | 0.23±0.06 | 0.14±0.05 | 0.17±0.05 |
| | | $E_{RMS}$ | 0.3±0.06 | 0.33±0.07 | 0.27±0.04 | 0.21±0.03 | 0.12±0.02 | 0.13±0.03 | 0.11±0.02 |
| | | $E_{L10}$ | 0.34±0.07 | 0.26±0.08 | 0.28±0.03 | 0.25±0.04 | 0.14±0.02 | 0.16±0.03 | 0.13±0.03 |
| | | %C | 0% | 0% | 0% | 0% | 0% | 0% | 20% |
| | G.M. (m/s^2) | $A_{RMS}$ | 0.55±0.29 | 0.39±0.17 | 0.40±0.18 | 0.27±0.18 | 0.16±0.05 | 0.08±0.03 | 0.08±0.02 |
| d1 | Lateral Error (m) | $E_{RNG}$ | 0.43±0.09 | 0.34±0.11 | 0.31±0.09 | 0.31±0.03 | 0.17±0.04 | 0.15±0.09 | 0.12±0.03 |
| | | $E_{RMS}$ | 0.39±0.04 | 0.31±0.06 | 0.27±0.04 | 0.29±0.02 | 0.16±0.03 | 0.14±0.03 | 0.11±0.03 |
| | | $E_{L10}$ | 0.43±0.07 | 0.27±0.06 | 0.18±0.04 | 0.23±0.02 | 0.22±0.04 | 0.13±0.05 | 0.12±0.04 |
| | | %C | 0% | 0% | 20% | 90% | 80% | 90% | 80% |
| | G.M. (m/s^2) | $A_{RMS}$ | 0.30±0.13 | 0.31±0.09 | 0.30±0.08 | 0.13±0.03 | 0.10±0.03 | 0.08±0.02 | 0.08±0.02 |
| e1 | Lateral Error (m) | $E_{RNG}$ | 0.47±0.09 | 0.27±0.07 | 0.21±0.1 | 0.36±0.06 | 0.29±0.02 | 0.13±0.05 | 0.14±0.04 |
| | | $E_{RMS}$ | 0.42±0.06 | 0.26±0.04 | 0.16±0.03 | 0.21±0.03 | 0.21±0.02 | 0.11±0.03 | 0.10±0.07 |
| | | $E_{L10}$ | 0.45±0.03 | 0.17±0.03 | 0.10±0.05 | 0.1±0.05 | 0.21±0.03 | 0.09±0.02 | 0.11±0.09 |
| | | %C | 40% | 50% | 60% | 0% | 80% | 90% | 80% |
| | G.M. (m/s^2) | $A_{RMS}$ | 0.32±0.12 | 0.27±0.07 | 0.19±0.13 | 0.23±0.11 | 0.18±0.12 | 0.08±0.01 | 0.06±0.02 |
| f1 | Lateral Error (m) | $E_{RNG}$ | 1.03±0.15 | 1.01±0.11 | 0.98±0.34 | 0.57±0.09 | 0.36±0.24 | 0.38±0.11 | 0.37±0.12 |
| | | $E_{RMS}$ | 0.24±0.03 | 0.22±0.1 | 0.21±0.03 | 0.36±0.02 | 0.08±0.05 | 0.10±0.04 | 0.12±0.06 |
| | | $E_{L10}$ | 0.07±0.03 | 0.05±0.03 | 0.04±0.04 | 0.30±0.04 | 0.06±0.05 | 0.06±0.02 | 0.02±0.03 |
| | | %C | 100% | 100% | 100% | 20% | 90% | 90% | 100% |
| | G.M. (m/s^2) | $A_{RMS}$ | 0.78±0.31 | 0.85±0.17 | 0.88±0.12 | 0.47±0.17 | 0.43±0.21 | 0.41±0.05 | 0.42±0.05 |

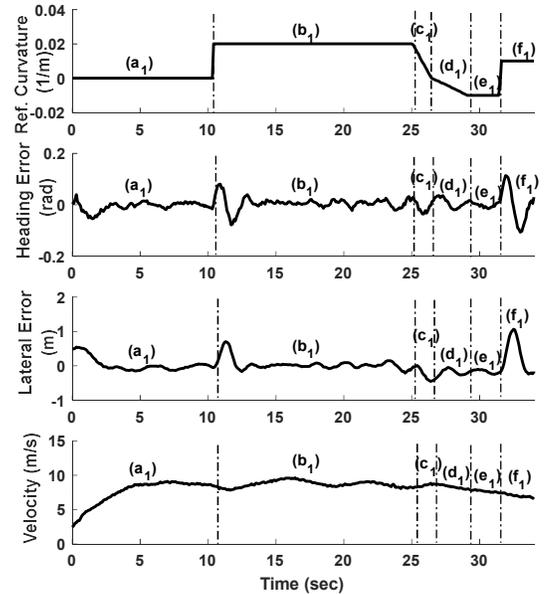

Fig 10: Experimental results: path curvature and tracking error of PROP-S using Comprehensive Path Plan 1 (8~10 $m/s$).

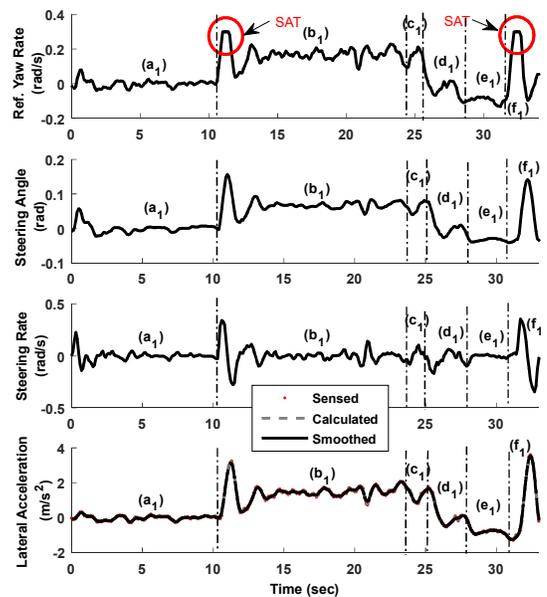

Fig 11: Experimental results: steering command and graceful motion index of PROP-S using Comprehensive Path Plan 1 (8~10 $m/s$).

Peaking saturation in PROP-S further benefits SEGs following engagement of the curvature saturation. SEG $b_1$ is followed by continuously varying curvature SEGs, $c_1$, $d_1$, and $e_1$. Both PROP and PROP-S demonstrate more graceful steering commands and lateral acceleration on continuous SEGS and do not engage the saturation. However, yaw rate commands saturated in nearly all PROP-S trials at the discontinuities. Performance is further improved using PROP-S on $c_1$, $d_1$, and $e_1$ even though saturation is rarely applied. *This is attributed to the fact that motion is smoother with PROP-S during $b_1$, which propagates through $c_1$ and $d_1$ until motion is improved in $e_1$.* While traversing $c_1$, $E_{RNG}$ and $E_{RMS}$ are 12% and 18% smaller. $E_{L10}$ is slightly larger, but standard deviations of $E_{RMS}$ and $E_{L10}$ are 43% and 63% smaller. Given this more consistent and improved tracking, $E_{RMS}$ and $E_{L10}$ are 31% and 38% smaller, respectively, on $d_1$. This propagates to $e_1$ where $E_{RMS}$, $E_{L10}$, and $A_{RMS}$ are 38%, 41%, and 30% smaller. Saturation only activated during one trial on $c_1$, and otherwise the controllers were identical on these trials. **While the peaking saturation of PROP-S may not always engage, its activation during earlier SEGs have a lasting effect.**

### B. Basic Path Studies – MEB Parking Lot

MEB results demonstrate the effect of perturbations caused by sloped terrain, **Table 3**. Clear weather conditions and dry surfaces are discussed first, followed by rainy conditions. **Results from PROP-S are not included since speeds were lower and yaw-rate saturation did engage, making PROP and PROP-S equivalent.**

The "L"-shaped path compares performance with straight (1st and 3rd SEG) and constant radius (2nd SEG) with large initial error and challenging curvature discontinuities between SEGs. PROP's $E_{RNG}$ is reduced ~5% to ~48% compared to B; $E_{RNG}$ is comparable on SEG 1, but greatly improved on SEGS 2 and 3. Compared to A, PROP reduced $E_{RNG}$ by ~20% to ~67%. $E_{RNG}$ is larger due to the initial posture error and curvature discontinuities, is smaller for PROP due to reduced overshoot and oscillation.

**PROP has improved accuracy, which is notable on the curved path (2nd SEG)** where $E_{RMS}$ is ~81% and ~79% less than A and B. PROP's final error, $E_{L10}$, is improved relative to B, although A is similar to PROP since the PID controller is suited to long constant curvature SEGs. This is not true on the 3rd SEG, which is shorter and straight, where A cannot fully converge, but B performs like PROP since the SEG is straight and PROP's sideslip compensation is less critical.

The "S" shaped path evaluates performance with an Euler spiral. PROP's overshoot is the best; $E_{RNG}$ is reduced ~41% compared to A and B. The first SEG has a large initial lateral error and ends with zero curvature, hence all three controllers have similar $E_{RMS}$, but PROP's $E_{L10}$ is reduced. **The 2nd SEG better indicates advantages of PROP since curvature progressively increases, which makes path following challenging.** PROP performs best; $E_{RMS}$ is ~45% to ~60% less than A and B. PROP provides better final error, $E_{L10}$; ~60% and ~77% less than A and B. **These improvements along the increasing curvature path SEG indicate the benefit of PROP following challenging paths.**

Convergence success rate $\%C$ indicates the ability to drive lateral error to a steady state value along a limited length path. PROP performed significantly better on the S-Shaped path compared to A and B. PROP provides $\%C = 40\%$ on the 1st SEG, while A and B provide 0%, which is attributed to the short segment and initial error. $\%C$ is improved for all controllers on the 2nd SEG due to smaller error at the start of the SEG. PROP ($\%C = 90\%$) is better than B ($\%C = 70\%$), but comparable to A. **Overall, PROP demonstrates notably improved convergence.**

PROP and B gracefulness is better than A. $A_{RMS}$ for PROP and B are generally ~50% less than A. PROP's $A_{RMS}$ is ~86% less on the "S"-shaped Euler spiral compared to the "L"-shaped path due to continuous curvature. **Improved gracefulness is attributed to the hierarchical path manifolds used in PROP and B,** although continuous paths further improve gracefulness, as expected.

Terrain geometry and surfaces impact convergence rate and tracking accuracy. MEB is difficult due to its slope and a dip in the parking lot. "A" has $\%C = 50\%$ on the 2nd SEG of the "L"-shaped path due to the low robustness of PID, while PROP and B provide $\%C = 100\%$ on this SEG. The 3rd SEG is a downhill straight path without lateral disturbances, so $\%C = 100\%$.

Rainy conditions were challenging for A and B convergence, $\%C$, but had small impact on PROP. PROP had a slight decrease in $\%C$, but A and B were significantly worse. While other metrics were similar in rain vs dry, this indicates that **PROP provides improved convergence despite weather conditions**.

Lateral acceleration, Fig 12, highlights graceful motion as characterized by $A_{RMS}$. Initial oscillations are due to initial conditions. "A" has large overshoot ($>0.5\ m/s^2$) and the most oscillation. "B" has less overshoot (~$0.5\ m/s^2$) and smoother lateral acceleration. PROP has the least oscillation and best follows the reference path acceleration. $A_{RMS}$ is essentially the same for PROP and B since they both use similar continuous VSC.

**Table 3: MEB experimental results using basic paths on clear and rainy days.**

| Basic Paths | | | MEB Parking Lot (10% Slope Asphalt Ground) | | | | | |
|---|---|---|---|---|---|---|---|---|
| | | | Clear Day | | | Rainy Day | | |
| | Performance Indices | | A | B | PROP | A | B | PROP |
| | | | avg ± σ | avg ± σ | avg ± σ | avg ± σ | avg ± σ | avg ± σ |
| L-shaped Path | Lateral Error | 1st Seg (m) | $E_{RNG}$ 0.82±0.05 | 0.62±0.05 | 0.65±0.08 | 0.81±0.16 | 0.60±0.05 | 0.59±0.05 |
| | | | $E_{RMS}$ 0.27±0.02 | 0.26±0.03 | 0.26±0.05 | 0.27±0.06 | 0.25±0.03 | 0.25±0.03 |
| | | | $E_{L10}$ 0.10±0.02 | 0.05±0.03 | 0.04±0.01 | 0.10±0.03 | 0.05±0.01 | 0.04±0.01 |
| | | | %C 100% | 100% | 100% | 60% | 60% | 100% |
| | | 2nd Seg (m) | $E_{RNG}$ 0.63±0.04 | 0.4±0.04 | 0.21±0.04 | 0.64±0.05 | 0.44±0.03 | 0.22±0.05 |
| | | | $E_{RMS}$ 0.26±0.01 | 0.24±0.03 | 0.05±0.01 | 0.26±0.01 | 0.28±0.03 | 0.05±0.02 |
| | | | $E_{L10}$ 0.05±0.02 | 0.26±0.05 | 0.03±0.02 | 0.04±0.02 | 0.34±0.04 | 0.04±0.03 |
| | | | %C 50% | 100% | 100% | 80% | 70% | 90% |
| | | 3rd Seg (m) | $E_{RNG}$ 0.59±0.05 | 0.33±0.05 | 0.24±0.04 | 0.68±0.04 | 0.37±0.07 | 0.32±0.08 |
| | | | $E_{RMS}$ 0.35±0.02 | 0.07±0.01 | 0.09±0.01 | 0.39±0.03 | 0.12±0.03 | 0.08±0.02 |
| | | | $E_{L10}$ 0.13±0.03 | 0.03±0.02 | 0.03±0.02 | 0.21±0.06 | 0.04±0.05 | 0.02±0.02 |
| | | | %C 100% | 100% | 100% | 100% | 100% | 100% |
| | G.M. | $A_{RMS}$ (m/s^2) | 1st Seg 0.22±0.02 | 0.14±0.03 | 0.13±0.03 | 0.26±0.06 | 0.13±0.03 | 0.10±0.02 |
| | | | 2nd Seg 0.22±0.02 | 0.20±0.03 | 0.18±0.02 | 0.25±0.02 | 0.21±0.03 | 0.17±0.05 |
| | | | 3rd Seg 0.28±0.04 | 0.23±0.03 | 0.22±0.03 | 0.31±0.05 | 0.29±0.03 | 0.31±0.06 |
| S-shaped : Euler spiral | Lateral Error | 1st Seg (m) | $E_{RNG}$ 0.75±0.14 | 0.61±0.09 | 0.63±0.05 | 0.70±0.05 | 0.62±0.11 | 0.64±0.07 |
| | | | $E_{RMS}$ 0.26±0.06 | 0.26±0.01 | 0.24±0.02 | 0.25±0.02 | 0.28±0.02 | 0.26±0.04 |
| | | | $E_{L10}$ 0.15±0.02 | 0.09±0.04 | 0.06±0.04 | 0.11±0.03 | 0.08±0.02 | 0.06±0.03 |
| | | | %C 0% | 0% | 40% | 0% | 0% | 30% |
| | | 2nd Seg (m) | $E_{RNG}$ 0.12±0.02 | 0.17±0.05 | 0.07±0.03 | 0.17±0.17 | 0.18±0.08 | 0.11±0.03 |
| | | | $E_{RMS}$ 0.11±0.03 | 0.15±0.02 | 0.06±0.04 | 0.17±0.07 | 0.16±0.02 | 0.10±0.02 |
| | | | $E_{L10}$ 0.10±0.04 | 0.19±0.01 | 0.04±0.03 | 0.11±0.06 | 0.21±0.03 | 0.09±0.03 |
| | | | %C 90% | 70% | 90% | 50% | 50% | 100% |
| | G.M. | $A_{RMS}$ (m/s^2) | 1st Seg 0.20±0.08 | 0.13±0.03 | 0.14±0.01 | 0.18±0.03 | 0.12±0.03 | 0.13±0.01 |
| | | | 2nd Seg 0.06±0.02 | 0.04±0.01 | 0.03±0.01 | 0.08±0.05 | 0.03±0.01 | 0.03±0.01 |

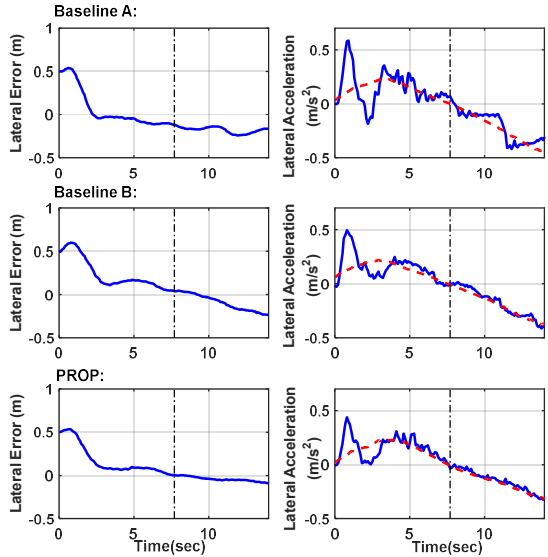

Fig 12: Steering controller comparison with the Euler spiral on MEB in rainy days. The 1st SEG occurs before the curvature inflection (e.g. at the vertical dash-dot line) and the 2nd SEG is after the inflection. The dashed line is from the reference path, solid is from experiments.

***PROP has the best overshoot $(< 0.5\ m/s^2)$, and its lateral acceleration fits the reference data well after settling.***

To help appreciate error convergence, Fig 12 compares lateral error and acceleration on the Euler spiral on MEB in rainy weather. Per **Table 3**, $\%C$ along the 1st SEG was the worst for all with failed convergence for A and B on this segment. Per Fig 12, A is clearly not converged at the curvature inflection, but B and PROP are less obvious. Both had increasing error just before the inflection and then error crossed zero at the last moment, so they were judged to not converge. Path length and weather conditions were major factors. ***The 2nd SEG occurs after the curvature inflection and highlights error convergence of the controllers.*** A and B do not converge, but PROP does since it remains near steady state. Lateral error with A oscillates around $\sim 0.2\ m$, highlighting difficulty with varying curvature. B diverges as curvature increases along the path. ***Again, PROP converges the best.***

### C. Basic Path Studies – SFP Parking Lot

Table 4 presents results from the SFP using basic paths on clear and rainy days. The SFP is a rough, uneven gravel surface. Results demonstrate similar trends as MEB, so this section highlights differences. With a few exceptions due to bumpy, uneven terrain, all controllers performed better on the SFP where sloped terrain disturbances were removed. PROP maintains better convergence, $\%C$, than A or B in most clear-day and rainy-day results. PROP has lower $\%C$ than in MEB (i.e., 70% vs. 100%) on the 1st SEG of "L"-shaped path in the rainy-day results. This illustrates that gravel in the SFP is challenging. "B" suffers slightly lower $\%C$ under the same situations (i.e., 50% vs. 60%) while A improved its convergence (i.e., 70% vs. 60%).

On the 2nd SEG, both PROP and B achieve 100% convergence, but A has 70%. This suggests that PROP and B can better handle curvature discontinuity even though the ground is uneven and slippery. Since the 2nd SEG is a longer arc, A has 90% convergence in SFP, which is better than $\%C = 50\%$ in MEB with the slope. On the 3rd SEG, PROP converges in all trials, but A and B have degraded performance compared to MEB. The SFP is bumpy and the 3rd SEG on MEB is mostly downhill, improving tracking.

A subtle performance degradation results from rainy conditions at the SFP. While numerical values are similar, error and steering commands tended to vary more. As a result, successful error convergence, $\%C$, was reduced for all controllers, although PROP was less adversely affected than A and B. One observation is that gracefulness is impacted more by terrain surfaces than slope disturbances, especially in the rainy days on the S-shaped path where $A_{RMS}$ raises 37.5%~166.7%.

## X. Discussion

Performance of the proposed controllers is highlighted by the field experiments in Sec. IX. The proposed controllers are first compared to the baseline controllers followed by discussion of control features exemplified by results.

### A. Comparison to Baseline Controllers

Experimental results demonstrate that PROP and PROP-S provide better performance than baseline controllers (i.e., A and B). $E_{RMS}$ and $E_{L10}$ values from the comprehensive path demonstrate that the proposed controllers are superior to the baseline controllers in complex paths. Unlike B, the proposed controllers have smaller $E_{RMS}$ and $E_{L10}$ on the constant and varying curvature SEGs, which is attributed to sideslip compensation. While A may achieve small $E_{L10}$ with long constant curvature SEGs, it does not on varying curvature SEGs due to the nature of PID control. Convergence success rate, $\%C$, demonstrates improved robustness of the proposed controllers on the comprehensive path. Speed and curvature variations are challenging, leading to varying sideslip. PROP and PROP-S had better $\%C$ because (54) and (61) allow them to consider steering actuator rate and provide robustness to lateral disturbances and uneven ground.

Tests on "rainy days" increased uncertainty due to slippery surfaces and related friction and cornering stiffness parameter variations. Performance on the "S" shaped Euler spiral highlights the benefit of PROP; the 2nd SEG is more challenging since curvature and slip are increasing. PROP has much smaller tracking error along the entire path. This is attributed to the features of the proposed control system discussed in the next sub section. Only minor performance differences are noted between clear and rainy days using PROP. On the "L"-shaped path, PROP has better convergence success, $\%C$, than the baseline controllers on all SEGs. While PROP shows slightly degraded $\%C$ in rainy conditions, it still achieves better success than baseline controllers.

Table 4: SFP experimental results using basic paths on clear and rainy days.

| Basic Paths | | | | SFP Parking Lot (Gravel Yard, Uneven Ground) | | | | | |
|---|---|---|---|---|---|---|---|---|---|
| | | | | Clear Day | | | Rainy Day | | |
| | | Performance Indices | | A | B | PROP | A | B | PROP |
| | | | | avg ± σ | avg ± σ | avg ± σ | avg ± σ | avg ± σ | avg ± σ |
| L-shaped Path | Lateral Error | 1st Seg (m) | $E_{RNG}$ | 0.70±0.09 | 0.53±0.08 | 0.50±0.03 | 0.71±0.08 | 0.54±0.04 | 0.51±0.06 |
| | | | $E_{RMS}$ | 0.25±0.04 | 0.25±0.02 | 0.28±0.01 | 0.23±0.03 | 0.26±0.03 | 0.26±0.05 |
| | | | $E_{L10}$ | 0.09±0.03 | 0.05±0.02 | 0.09±0.03 | 0.07±0.03 | 0.05±0.02 | 0.05±0.03 |
| | | | $\%C$ | 100% | 100% | 100% | 70% | 50% | 70% |
| | | 2nd Seg (m) | $E_{RNG}$ | 0.66±0.14 | 0.54±0.12 | 0.37±0.06 | 0.66±0.08 | 0.38±0.08 | 0.35±0.06 |
| | | | $E_{RMS}$ | 0.25±0.02 | 0.25±0.04 | 0.14±0.03 | 0.24±0.01 | 0.25±0.03 | 0.09±0.02 |
| | | | $E_{L10}$ | 0.06±0.04 | 0.24±0.08 | 0.05±0.03 | 0.05±0.03 | 0.25±0.04 | 0.03±0.02 |
| | | | $\%C$ | 90% | 100% | 100% | 70% | 100% | 100% |
| | | 3rd Seg (m) | $E_{RNG}$ | 0.55±0.05 | 0.36±0.09 | 0.37±0.03 | 0.68±0.10 | 0.32±0.08 | 0.34±0.05 |
| | | | $E_{RMS}$ | 0.37±0.02 | 0.10±0.08 | 0.11±0.02 | 0.36±0.04 | 0.13±0.04 | 0.12±0.04 |
| | | | $E_{L10}$ | 0.20±0.04 | 0.05±0.03 | 0.03±0.02 | 0.19±0.06 | 0.05±0.06 | 0.03±0.02 |
| | | | $\%C$ | 60% | 80% | 100% | 80% | 100% | 100% |
| | G.M. | $A_{RMS}$ (m/s^2) | 1st Seg | 0.23±0.05 | 0.11±0.03 | 0.11±0.02 | 0.22±0.05 | 0.11±0.03 | 0.11±0.03 |
| | | | 2nd Seg | 0.35±0.04 | 0.27±0.04 | 0.21±0.09 | 0.28±0.04 | 0.25±0.06 | 0.27±0.04 |
| | | | 3rd Seg | 0.28±0.05 | 0.24±0.03 | 0.25±0.05 | 0.46±0.19 | 0.32±0.03 | 0.27±0.08 |
| S-shaped : Euler spiral | Lateral Error | 1st Seg (m) | $E_{RNG}$ | 0.78±0.07 | 0.65±0.08 | 0.66±0.12 | 0.70±0.11 | 0.66±0.04 | 0.57±0.07 |
| | | | $E_{RMS}$ | 0.27±0.03 | 0.28±0.04 | 0.27±0.06 | 0.23±0.03 | 0.24±0.04 | 0.26±0.02 |
| | | | $E_{L10}$ | 0.14±0.02 | 0.03±0.01 | 0.04±0.01 | 0.10±0.02 | 0.03±0.02 | 0.04±0.01 |
| | | | $\%C$ | 40% | 40% | 80% | 0% | 10% | 80% |
| | | 2nd Seg (m) | $E_{RNG}$ | 0.17±0.04 | 0.22±0.04 | 0.13±0.03 | 0.12±0.01 | 0.19±0.04 | 0.12±0.03 |
| | | | $E_{RMS}$ | 0.15±0.02 | 0.14±0.02 | 0.08±0.03 | 0.14±0.01 | 0.12±0.01 | 0.11±0.02 |
| | | | $E_{L10}$ | 0.13±0.02 | 0.18±0.04 | 0.08±0.05 | 0.13±0.02 | 0.17±0.02 | 0.09±0.02 |
| | | | $\%C$ | 90% | 90% | 100% | 60% | 70% | 80% |
| | G.M. | $A_{RMS}$ (m/s^2) | 1st Seg | 0.26±0.05 | 0.13±0.04 | 0.13±0.01 | 0.19±0.03 | 0.15±0.04 | 0.14±0.03 |
| | | | 2nd Seg | 0.12±0.05 | 0.06±0.01 | 0.06±0.02 | 0.11±0.02 | 0.08±0.02 | 0.07±0.02 |

Similar trends occur with the Basic Paths where additional challenges are indicated. The "S"-shaped path is more challenging than the "L"-shaped path on the MEB sloped surfaces. Due to perturbations of varying curvature, gravity disturbance, and initial lateral error, PROP only has moderate $\%C$ along the 1st SEG, whereas Baseline controllers fail to converge initial error. $E_{RNG}$ and $A_{RMS}$ evaluate gracefulness. $E_{RNG}$ shows overshoot of tracking error and $A_{RMS}$ indicates oscillations and deviation from ideal lateral acceleration. On the SFP comprehensive path, PROP and PROP-S suffer less overshoot and oscillation than baseline controllers and present lower $E_{RNG}$ and $A_{RMS}$ at higher speeds.

### B. Controller Features

Superior performance of PROP and PROP-S compared to A and B is attributed to the novel control structure, slip-yaw kinematic and dynamic models, and related sideslip feedback compensation. PROP and PROP-S apply a closed-loop control structure that feeds back sideslip estimates to compensate heading error, which is generally missed in A and B.

PROP and PROP-S apply additional features to resolve issues in tracking accuracy, convergence success, and graceful motion. The architecture separates the steering controller into four tracking controllers and distributes them in the closed-loop control structure. Sideslip compensation allows the kinematic controller to continually adapt to increasing sideslip. The robust VSC kinematic controller naturally considers model-based terms and compensates for uncertainty via $\psi_{kin}$, while continuous implementation of the controller eliminates chatter. Varying $c(t)$ adapts to increasing speed and reduces oscillation during initial error convergence. Thus, the controllers can reject uncertainty and rapidly stabilize tracking error to achieve convergence before a SEG ends.

In the dynamic controller, backstepping is critical for designing steering actuator rate commands that consider the cumulative requirements of kinematic and dynamic controllers.

Integrators distributed in the control architecture allow remaining perturbations in kinematics, dynamics, and steering actuation to be compensated while avoiding large control gains.

Although PROP and PROP-S are essentially the same design, saturation in PROP-S plays a critical role to reducing "peaking" at curvature discontinuities at high speeds. Thus, $E_{RNG}$ and $A_{RMS}$ are further decreased in PROP-S compared to PROP.

### C. Future work

There are a number of topics for future work. First, instead of using fixed yaw rate saturation and manually tuning it, automatic saturation adaption should be examined. Factors such as yaw rate command variation, vehicle speed, path complexity, and environmental constraints should be considered. Coupling between lateral and longitudinal dynamics should be considered further. Although speed variations are currently treated as a disturbance, a fully coupled lateral and longitudinal dynamic model should be considered to reduce their effect on the steering controller and improve response. While the proposed controllers performed well with uncertain parameters, online parameter identification should reduce dependence on robust terms and improve gracefulness. Future work could also focus on combining the controller with sensor-based navigation, but paths should have continuous curvature variations. This highlights that rates of curvature variation need to be considered, which is related to the saturation levels that are used in PROP-S.

## XI. CONCLUSION

A comprehensive approach considering kinematics, dynamics, actuation, and state estimation in steering control is proven to provide good path following accuracy. Slip is important to consider in kinematic models and controllers. Path manifolds need to adapt to operating conditions. Tracking accuracy should be considered at all levels. Backstepping is valuable for mapping all these factors to steering actuator rate commands. This work relies upon the application of a slip-yaw dynamic model, yaw reference commands from the kinematic controller, and high-gain observer-based estimation of slip and yaw states. Rigorous field experiments demonstrate improved performance of the proposed system compared to baseline studies. Multiple test fields with different surfaces and terrain disturbances demonstrate that the proposed controller provides improved tracking accuracy, robustness, and graceful motion, which is especially noteworthy with difficult real-world time-varying paths, environmental disturbances, and adverse weather conditions. Peaking should be considered at higher speeds. Future work should emphasize online adaptation of controller parameters and identification of vehicle parameters.

## APPENDIX

*Proof of Theorem 1:* Consider (27) in the unsaturated domain $D_1$ where $S_{kin} = \bar{\theta}_e + \arcsin\left(\frac{c(t)y_e + K_i\sigma_k}{\bar{v}}\right)$, which is continuously differentiable.

The positive definite Lyapunov candidate function $W_{kin} = \frac{1}{2}S_{kin}^2$ is applied such that $\dot{W}_{kin} = S_{kin}\dot{S}_{kin}$, where $\dot{S}_{kin}$ is:

$$\dot{S}_{kin} = \dot{\bar{\theta}}_e + \frac{\dot{c}(t)y_e + c(t)(\bar{v}\sin\bar{\theta}_e - \bar{v}\delta_{\alpha r}) + K_i y_e}{\bar{v}\sqrt{1-\left(\frac{c(t)y_e + K_i\sigma_k}{\bar{v}}\right)^2}}. \quad (62)$$

The time derivative of (21) is then $\dot{\bar{\theta}}_e = \dot{\theta}_e + K_F\dot{\hat{\beta}}$, which is combined with (13) to form the state equation for $\bar{\theta}_e$,

$$\dot{\bar{\theta}}_e = \kappa_{ref}v_{ref} - r_{kin} + K_F\dot{\hat{\beta}}, \quad (63)$$

where $r$ in (13) has been replaced by $r_{kin}$ to denote the kinematic control command. (63) is then combined with (62),

$$\dot{S}_{kin} = \kappa_{ref}v_{ref} - r_{kin} + K_F\dot{\hat{\beta}} + \frac{\dot{c}(t)y_e + c(t)(\bar{v}\sin\bar{\theta}_e - \bar{v}\delta_{\alpha r}) + K_i y_e}{\bar{v}\sqrt{1-\left(\frac{c(t)y_e + K_i\sigma_k}{\bar{v}}\right)^2}} \quad (64)$$

and $\dot{W}_{kin} = S_{kin}\dot{S}_{kin} =$

$$= S_{kin}(\kappa_{ref}v_{ref} - r_{kin} + K_F\dot{\hat{\beta}} + \frac{\dot{c}(t)y_e + c(t)(\bar{v}\sin\bar{\theta}_e - \bar{v}\delta_{\alpha r}) + K_i y_e}{\bar{v}\sqrt{1-\left(\frac{c(t)y_e + K_i\sigma_k}{\bar{v}}\right)^2}}). \quad (65)$$

Rearranging terms for the VSC controller proof,

$$\dot{W}_{kin} = S_{kin}(\kappa_{ref}v_{ref} + K_F\dot{\hat{\beta}} + \frac{\dot{c}(t)y_e + c(t)(\bar{v}\sin\bar{\theta}_e - \bar{v}\delta_{\alpha r}) + K_i y_e}{\bar{v}\sqrt{1-\left(\frac{c(t)y_e + K_i\sigma_k}{\bar{v}}\right)^2}}) - r_{kin}S_{kin} \quad (66)$$

which can be combined with (30) to prove that,
$$\dot{W}_{kin} \leq |S_{kin}|\rho_{kin} + S_{kin}(\kappa_{ref}v_{ref} + K_F\dot{\hat{\beta}}) - r_{kin}S_{kin}. \quad (67)$$
Applying the designed yaw rate command, $r_{kin}$,
$$r_{kin} = \kappa_{ref}v_{ref} + K_F\dot{\hat{\beta}} + (\rho_{kin} + \psi_{kin})\tanh\left(\frac{S_{kin}}{\varepsilon_{kin}}\right) \quad (68)$$
into (67), such that:
$$\dot{W}_{kin} \leq -(\rho_{kin} + \psi_{kin})\tanh\left(\frac{S_{kin}}{\varepsilon_{kin}}\right)S_{kin} + |S_{kin}|\rho_{kin} \quad (69)$$
The effect of tanh() [20] is considered by noting that,
$$\begin{cases} \left|\tanh\left(\frac{S_{kin}}{\varepsilon_{kin}}\right)\right| \cong 1 & , if \left|\frac{S_{kin}}{\varepsilon_{kin}}\right| \geq \alpha_0 \quad (a) \\ 0 \leq \left|\tanh\left(\frac{S_{kin}}{\varepsilon_{kin}}\right)\right| < 1 & , if \left|\frac{S_{kin}}{\varepsilon_{kin}}\right| < \alpha_0 \quad (b) \end{cases} \quad (70)$$
where $\alpha_0$ is the bound value and $\varepsilon_{kin}$ is tunable.

In condition (a) in (70), i.e. $\tanh\left(\frac{S_{kin}}{\varepsilon_{kin}}\right)$ is saturated, $\tanh\left(\frac{S_{kin}}{\varepsilon_{kin}}\right)S_{kin} \equiv \text{sign}(S_{kin})S_{kin} = |S_{kin}|$. Thus, (69) becomes $\dot{W}_{kin} \leq -\psi_{kin}|S_{kin}| < 0$ for all $S_{kin} \neq 0$ such that $\dot{W}_{kin}$ is negative definite. Trajectories converge to the boundary $|S_{kin}| = \alpha\varepsilon_{kin}$ in finite time [45]. Due to $\dot{W}_{kin} \leq -\psi_{kin}|S_{kin}| < 0$ on the boundary, trajectories are then directed inwards and cannot leave.

Once inside $|S_{kin}| = \alpha\varepsilon_{kin}$, condition (b) is satisfied and $\tanh\left(\frac{S_{kin}}{\varepsilon_{kin}}\right)$ stays in the unsaturated domain. The tanh() term is represented by $\tanh\left(\frac{S_{kin}}{\varepsilon_{kin}}\right) = \zeta$, which is expressed as $\zeta = |\zeta|\text{sign}\left(\frac{S_{kin}}{\varepsilon_{kin}}\right) = |\zeta|\text{sign}(S_{kin})$ due to $\varepsilon_{kin} > 0$, where $0 \leq \zeta < 1$. Substituting into (69), we have $\dot{W}_{kin} \leq -(\rho_{kin} + \psi_{kin})|\zeta|\text{sign}(S_{kin})S_{kin} + |S_{kin}|\rho_{kin}$. Thus,
$$\dot{W}_{kin} \leq -(\psi_{kin}|\zeta| - \rho_{kin}(1 - |\zeta|))|S_{kin}| \quad (71)$$
Since $\psi_{kin}$ is an arbitrary positive number, it can always be selected to ensure that $\psi_{kin}|\zeta| \geq \rho_{kin}(1 - |\zeta|)$ and results in $\dot{W}_{kin} < 0$ where $S_{kin} \neq 0$. Thus, $\dot{W}_{kin}$ is always negative definite. Since the right hand side is multiplied by $|S_{kin}|$, it can be shown that trajectories reach $S_{kin} = 0$ in finite time and cannot leave the path manifold [45]. Fig 13 (top) verifies variation of $\dot{W}_{kin}$ vs. $S_{kin}$ with a practical parameter set, $\psi_{kin} = 0.1$, $\varepsilon_{kin} = 0.1$, $\bar{v} = 10 m/s$, $K_i = 0.04$, $\dot{c}(t) = 0$, $c(t) = c_{final} = 0.65$. Per the parameters, $\psi_{kin}|\zeta| - \rho_{kin}(1 - |\zeta|) > 0$ for all $S_{kin} \neq 0$, such that (71) succeeds. Once on the path manifold, Proposition 4 is applied to show that the net result is $x_2 \to 0$ asymptotically.

Fig 13 (bottom) presents the linearized approximation of $\tan\left(\frac{S_{kin}}{\varepsilon_{kin}}\right)$ where the intersections with dash lines illustrate $|\frac{S_{kin}}{\varepsilon_{kin}}| \geq |\zeta| \geq |\frac{S_{kin}}{k\varepsilon_{kin}}|$, where $k = 2$ or $3$ show lower bounds on the function based upon 95% or 99% of saturated values, respectively, at the transition. As it can be seen, the $tanh()$ function is bounded above and below by linear approximations.

Local exponential stability can be proven by applying the controller (29) to the model (26), (28), (13), resulting in,
$$\begin{aligned} \dot{\sigma}_k &= y_e \\ \dot{y}_e &= \bar{v}\sin\bar{\theta}_e \\ \dot{\bar{\theta}}_e &= -(\rho_{kin} + \psi_{kin})\tanh\left(\frac{S_{kin}}{\varepsilon_{kin}}\right) \end{aligned} \quad (72)$$
The Jacobian matrix at the equilibrium $x_{2ss} = [0 \quad 0 \quad \sigma_{ss}]^T$ is,
$$J_{x_2} = \begin{bmatrix} 0 & 1 & 0 \\ 0 & 0 & \bar{v} \\ -\left(\frac{\psi_{kin}K_i}{\bar{v}\varepsilon_{kin}}\right) & \pm\frac{(K_i+\dot{c}(t))K_i\sigma_{ss}}{\bar{v}^2\sqrt{1-\frac{K_i^2\sigma_{ss}^2}{\bar{v}^2}}\varepsilon_{kin}} - \frac{\psi_{kin}c(t)}{\bar{v}\varepsilon_{kin}} & \pm\frac{c(t)K_i\sigma_{ss}}{\bar{v}\varepsilon_{kin}\sqrt{1-\frac{K_i^2\sigma_{ss}^2}{\bar{v}^2}}} - \frac{\psi_{kin}}{\varepsilon_{kin}} \end{bmatrix} \quad (73)$$
Substituting the above parameter configuration into (73) and using the upper bound on tanh() which is more accurate near the origin, the eigenvalues are $\lambda(J_{x_2}) = -0.068, -0.466 \pm 0.608i$. Since $Re(\lambda(J_{x_2})) < 0$, $J_{x_2}$ is Hurwitz, such that $x_2$ in the unsaturated domain is exponentially stable. The lower bounds on tanh() can be applied to show exponential stability as the function enters the saturation region, but with a slower response.

Finally, $\dot{\hat{\beta}}$ in (68) is noisy in application and results in noisy $r_{kin}$. If $\dot{\hat{\beta}}$ is included, noise will propagate to the dynamic controller where the problem is worsened by the appearance of its 2nd and 3rd time derivatives, causing poor performance. Hence, $K_F\dot{\hat{\beta}}$ is ignored. Thus, (68) becomes (29) per the theorem.∎

Proof of Theorem 5: The integral of steering angle error is calculated by $\dot{\sigma}_\varphi = \varphi_e$. The time derivative of $\varphi_e$ provides $\dot{\varphi}_e = \dot{\varphi}_{des} - \dot{\varphi}_{act}$. The time derivative of (40) replaces $\dot{\varphi}_{des}$ whereas $\dot{\varphi}_{act}$ is based upon the vehicle dynamics (16) where $\varphi$ is denoted as $\varphi_{act}$ and thus $\dot{\varphi}_{act} = \omega$, the steering rate. The result is,
$$\dot{\varphi}_e = \dot{\varphi}_{des} - \dot{\varphi}_{act} = -\frac{a_{21}\dot{\beta}+a_{22}\dot{r}_{kin}-\ddot{r}_{kin}-K_{p1}\dot{r}_e-K_{i1}r_e}{b_{21}} - \omega . \quad (74)$$
Time invariant backstepping is applied since time-varying terms vary slowly. The composite Lyapunov candidate function is,
$$W_c = \frac{1}{2}r_e^2 + \frac{K_{i1}}{2}\sigma_r^2 + \frac{1}{2}\varphi_e^2 + \frac{K_{i2}}{2}\sigma_\varphi^2 \quad (75)$$
where $W_c: D_c \to \Re$ is a continuously differentiable positive definite function on a domain, $D_c$, containing an equilibrium point $x_c \equiv x_{c_{ss}}$, where $x_c = [r_e \quad \sigma_r \quad \varphi_e \quad \sigma_\varphi]^T$. $\dot{W}_c$ is then:
$$\dot{W}_c = r_e\dot{r}_e + K_{i1}\sigma_r\dot{\sigma}_r + \varphi_e\dot{\varphi}_e + K_{i2}\sigma_\varphi\dot{\sigma}_\varphi \quad (76)$$
Substituting the integrals and (74) into (76) and rearranging:
$$\dot{W}_c = -(K_{p1} + a_{22})r_e^2 + \varphi_e(\dot{\varphi}_e + K_{i2}\sigma_\varphi + r_e) \quad (77)$$
Since $K_{p1} > a_{22}$, $-(K_{p1} + a_{22})r_e^2 \leq 0$. $\omega$ is designed such that $\varphi_e(\dot{\varphi}_e + K_{i2}\sigma_\varphi + r_e) = -K_{p2}\varphi_e^2$. Hence, (77) becomes:
$$\dot{W}_c = -(K_{p1} + a_{22})r_e^2 - K_{p2}\varphi_e^2 \quad (78)$$
and results in $\dot{\varphi}_e + K_{i2}\sigma_\varphi + r_e = -K_{p2}\varphi_e$, such that:
$$-\frac{a_{21}\dot{\beta}+a_{22}\dot{r}_{kin}-\ddot{r}_{kin}-K_{p1}\dot{r}_e-K_{i1}r_e}{b_{21}} - \omega + K_{i2}\sigma_\varphi + r_e = -K_{p2}\varphi_e \quad (79)$$
Rearranging (79), the steering actuator rate $\omega$ is shown in (42).

Given $K_{p1}$ and $K_{p2}$ above, $\dot{W}_c \leq 0$ per (78), which is negative semi-definite. Invariance is applied to prove asymptotic stability. A domain, $S_c$, is set as $S_c = \{x_c \in D_c | \dot{W}_c(x_c) = 0\}$, where $W_c: D_c \to \Re$ contains the origin $x_c \equiv x_{c_{ss}}$. The domain $D_c$ is defined as $D_c = \{x_c \in \Re^4 | 0 < r_e^2 + K_{i1}\sigma_r^2 + \varphi_e^2 + K_{i2}\sigma_\varphi^2 < a_c, a_c \in \Re\}$. No solution can stay identically in $S_c$, other than the trivial solution $x_c \equiv x_{c_{ss}}$. Based on Barbashin's theorem [45], the equilibrium point $x_{c_{ss}}$ is asymptotically stable. Note that $\dot{W}_c(x_c) = 0$ and (78) guarantees that $r_e$ and $\varphi_e$ converge to zero, although integrator states may converge to nonzero values. Thus, (42) is the dynamic control law. ∎

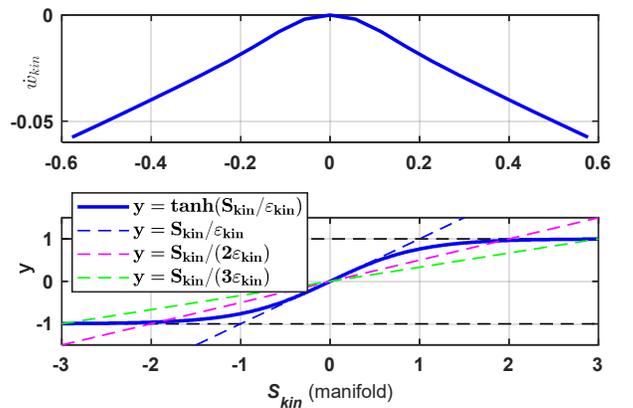

Fig 13: Variation of the Hyperbolic Tangent Function (tanh ($S_{kin}/\varepsilon_{kin}$)) and Its Linearized Approximation at different gains

BIOGRAPHIES

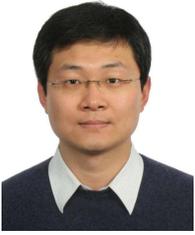

**Ming Xin** received the M. S. degree in Robotics and Controls from the mechanical engineering department in Ohio University, and the Ph. D. degree in Robotics from the Mechanical Engineering Department in University of Utah, Salt Lake City, Utah, USA. He is currently with Xpeng as director of control systems. His research interests focus on robot controls, path planning, system estimation, and parameter identification.

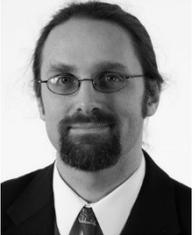

**Mark A. Minor** (M00) received the B.S. degree in mechanical engineering from the University of Michigan, Ann Arbor, in 1993, and the M.S. and Ph.D. degrees in mechanical engineering from Michigan State University, East Lansing, in 1996 and 2000, respectively. He is currently an Associate Professor in Mechanical Engineering, University of Utah, Salt Lake City, where he has been a faculty member since 2000. His research interests focus on design and control of robotic systems including mobile robots, rolling robots, climbing robots, flying robots, autonomous ground vehicles, soft robots, wearable robots, and virtual reality systems.